\documentclass[journal]{IEEEtran}
\usepackage{cite}
\usepackage{amsmath,amssymb,amsfonts}
\usepackage{graphicx}
\usepackage{textcomp}
\usepackage{xcolor}
\usepackage{algorithm}
\usepackage{algorithmic}
\usepackage{multirow}
\usepackage{enumerate}
\usepackage{url}
\usepackage{booktabs}
\usepackage{threeparttable}
\usepackage{endnotes}
\usepackage{subfigure}
\usepackage{comment}
\usepackage[switch,pagewise,columnwise]{lineno}
\usepackage[colorlinks=true,linkcolor=blue,citecolor=blue,urlcolor=blue]{hyperref}

\ifCLASSINFOpdf
\else
\fi

\begin{document}
%
\title{Frequent Itemset-driven Search for Finding Minimum Node Separators in Complex Networks}
\author{Yangming~Zhou,
        Xiaze~Zhang,
        Na~Geng,
        Zhibin~Jiang
        and Mengchu~Zhou
\thanks{This work was supported in part by the National Natural Science Foundation of China under Grant No. 61903144 and No. 72031007; and in part by the Macau Young Scholars Program under Grant No. AM2020011. (\emph{Corresponding author: Mengchu~Zhou})}
\thanks{Yangming~Zhou is with the Sino-US Global Logistics Institute, Shanghai Jiao Tong University, Shanghai 200030, China, and also with the Macau Institute of System Engineering, Macau University of Science and Technology, Macau 999078, China (e-mail: yangming.zhou@sjtu.edu.cn).}
\thanks{Xiaze~Zhang is with the Department of Computer Science and Engineering, East China University of Science and Technology, Shanghai 200237, China (e-mail: zhangxiaze@gmail.com).}
\thanks{Na~Geng is with the Sino-US Global Logistics Institute, Shanghai Jiao Tong University, Shanghai 200030, China (email: gengna@sjtu.edu.cn).}
\thanks{Zhibin~Jiang is with the Antai College of Economics and Management, Shanghai Jiao Tong University, Shanghai 200030, China (email: zbjiang@sjtu.edu.cn).}
\thanks{Mengchu~Zhou is with the Department of Electrical and Computer Engineering, New Jersey Institute of Technology, Newark, NJ 07102, USA (e-mail: zhou@njit.edu).}
}


\maketitle
\begin{abstract}

Finding an optimal set of critical nodes in a complex network has been a long-standing problem in the fields of both artificial intelligence and operations research. Potential applications include epidemic control, network security, carbon emission monitoring, emergence response, drug design, and vulnerability assessment. In this work, we consider the problem of finding a minimal node separator whose removal separates a graph into multiple different connected components with fewer than a limited number of vertices in each component. To solve it, we propose a frequent itemset-driven search approach, which integrates the concept of frequent itemset mining in data mining into the well-known memetic search framework. Starting from a high-quality population built by the solution construction and population repair procedures, it iteratively employs the frequent itemset recombination operator (to generate promising offspring solution based on itemsets that frequently occur in high-quality solutions), tabu search-based simulated annealing (to find high-quality local optima), population repair procedure (to modify the population), and rank-based population management strategy (to guarantee a healthy population). Extensive evaluations on 50 widely used benchmark instances show that it significantly outperforms state-of-the-art algorithms. In particular, it discovers 29 new upper bounds and matches 18 previous best-known bounds. Finally, experimental analyses are performed to confirm the effectiveness of key algorithmic modules of the proposed method.

\end{abstract}

\begin{IEEEkeywords}
Metaheuristics; Memetic search; Data mining; Critical node detection; $\alpha$-separator problem
\end{IEEEkeywords}

%
\IEEEpeerreviewmaketitle

\section{Introduction}
\label{Sec:Introduction}

A network is a kind of widely used data structure to describe numerous types of interactive systems, such as a supply chain system, communication network, smart grid, transportation system, biological network and social network. It is usually affected by a small fraction of important nodes whose activation (or removal) would greatly enhance (or degrade) network functionality. Depending on their roles in different application scenarios, such nodes have been named differently, such as critical nodes \cite{Naoum-Sawaya2016,Baggio2021,Zhou2021a}, key players \cite{Borgatti2006,Fan2020}, influential nodes \cite{Morone2015,Kamarthi2020,Wang2021a}, and critical links \cite{Veremyev2019}. Detecting these important nodes are known as critical node detection problems (CNDPs) \cite{Arulselvan2009,Lalou2018,Zhou2021a,Baggio2021}. CNDPs deal with identifying a set of nodes from a network, whose deletion optimizes a predefined network connectivity measure over the residual network, which are typically NP-hard \cite{Arulselvan2009}.

Finding an optimal set of critical nodes in a complex network is a fundamental class of problems in network science. A number of real-world applications can be naturally modelled as CNDPs, i.e., 1) epidemic control \cite{Ghalmane2019,Doostmohammadian2020}, where the goal is to identify only a specific number of people to be vaccinated in order to reduce the overall transmissibility of a virus; 2) network security \cite{Mugisha2016}, where the attacker is interested in disabling some most important nodes to make the network more vulnerable, while the defender is interested in reinforcing the protection over these important nodes and applying more robust security measures; 3) carbon emission monitoring \cite{Zhang2020b,Zhao2021,Wen2021}, where the target is to find the critical paths and nodes that contribute strongly to carbon emissions embodied in transmission; 4) emergence response \cite{Vitoriano2011}, which attempts to identify some critical nodes that can be used to plan good emergency evacuations in a disaster cases; 5) drug design \cite{Tomaino2012}, which aims to destroy certain critical proteins and neutralize the corresponding harmful protein complexes for rational drug design; and 6) social network analysis \cite{More2019} whose main objective is to find the most influential entities within the social network.

CNDPs can be divided into two categories: $k$-vertex-CNDP and $\beta$-connectivity-CNDP. The former aims to optimize the connectivity metric $\sigma$, such that no more than $k$ nodes are deleted; while the latter is to minimize the set of deleted nodes such that $\sigma$ is bound by a given threshold $\beta$. Due to their theoretical and practical significance, CNDPs have attracted considerable efforts \cite{Arulselvan2009,Lalou2018,Zhou2019,Zhou2021a,Baggio2021,Wu2021b}. However, most of existing studies are devoted to solving the $k$-vertex-CNDPs, such as critical node problem (CNP) \cite{Arulselvan2009,Pullan2015,Zhou2019,Zhou2021a,Baggio2021}. Few have been made on developing efficient algorithms for $\beta$-connectivity-CNDP. This work focuses on the $\alpha$-separator problem ($\alpha$-SP), which is a classic $\beta$-connectivity-CNDP. It consists in finding a minimum node separator (i.e., a set of nodes) $S \subseteq V$ such that any connected component in the residual graph $G[V \setminus S]$ contains at most $\lceil \alpha \cdot n \rceil$ nodes, where $\alpha$ ($1/n \leq \alpha < 1$) is a pre-set parameter. For some special networks (e.g., trees and cycles), $\alpha$-SP can be solved in polynomial time \cite{Mohamed2014}. However, for a general topology network, it is NP-hard when $\alpha \leq 2/3$ \cite{Feige2006}.

$\alpha$-SP is closely related with other well-established combinatorial optimization problems. In particular, if $\alpha = 1/n$, it is equivalent to the classical vertex cover problem \cite{Katzmann2017,Li2018}. If $\alpha = 2/n$, it is analogous to a dissociation set problem \cite{Yannakakis1981}. Both problems were proven to be NP-hard in \cite{Garey1979}. When a larger $\alpha$ value is given, e.g., $\alpha = k/n$, it reduces to a $k$-separator problem \cite{Ben-Ameur2013}. $k$-separator problem has also been studied as a cardinality-constrained critical node problem \cite{Arulselvan2011}, which is a cardinality constrained version of CNP \cite{Zhou2019,Zhou2021a}. Therefore, $\alpha$-SP can be considered as a generalization of four optimization problems mentioned above. In addition, it is proven to be a useful model to cope with a variety of practical applications, e.g., network security that aims at deleting a minimum number of target nodes to break the network down into many small ones \cite{Mugisha2016}.

In this paper, we present an efficient Frequent Itemset-driven Search (FIS) method for $\alpha$-SP. It integrates the concept of a frequent itemset into a general memetic search framework. It is a well-established fact that generally many high-quality local optima of a combinational optimization problem share some common items. These common items are useful to guide the search. However, it is time-consuming to mine them from high-quality solutions by directly using existing data mining algorithms. This work attempts to make the following new contributions:
\begin{itemize}
    \item It develops an iterative solution strategy that considers $\alpha$-SP from the viewpoint of constraint satisfaction by solving a series of $K$-decision $\alpha$-SPs. To speed up the solution evaluation of $K$-decision $\alpha$-SP, it proposes a useful auxiliary function that counts the number of nodes in excess of $\lceil \alpha \cdot n \rceil$ in the largest connected component instead of all connected components.
    \item It presents an FIS method to solve $\alpha$-SP, where common itemsets shared among high-quality solutions are quickly identified based on only item frequency counts and then used to guide the search. It consists of five specifically designed and original modules: 1) a solution construction procedure to generate high-quality feasible solutions; 2) a population repair procedure to generate an initial population based on high-quality feasible solutions; 3) a frequent itemset recombination operator to quickly construct offspring solutions based on the frequent itemset shared by high-quality solutions; 4) tabu search-based simulated annealing to perform local optimization that relies on a two-phase node exchange strategy and a tabu search strategy; and 5) a rank-based population management strategy to maintain a healthy population by considering both candidate solution quality and solution distance.
    \item It performs extensive experiments to evaluate the performance of FIS and its state-of-the-art peers on 50 widely used benchmark instances. Experimental results show that FIS is able to find new upper bounds for 29 out of 50 tested instances and match pervious best-known bounds on 18 instances. It also shows its superiority over its peers in terms of both best and average results.
\end{itemize}

The rest of this paper is organized as follows: Section \ref{Sec:Problem Description and Related Work} gives the problem description of $\alpha$-SP and briefly reviews previous studies. Section \ref{Sec:Frequent Itemset Driven Search} presents an FIS approach for $\alpha$-separator problems. Section \ref{Sec:Empirical Results} conducts its experimental evaluations and comparisons with its peers. Section \ref{Sec:Experimental Result Analysis} presents additional experimental analyses of FIS. Finally, Section \ref{Sec:Conclusion} summarizes the work and gives potential directions.

\section{Problem Description and Related Work}
\label{Sec:Problem Description and Related Work}

\subsection{Problem Description}
\label{SubSec:Problem Description}

A network is usually described as a graph $G=(V,E)$, where $V$ is a vertex set ($|V|=n$) and $E$ is a edge set $|E|=m$. A vertex $v \in V$ represents a network node, while an edge $e(u,v) \in E, u,v \in V$ denotes a connection or link in the network between nodes $u$ and $v$, which are called endpoints of edge $e(u,v)$. For a vertex $v$, its neighborhood is $N(v)=\{u \in V|e(u,v)\in E\}$, and its degree is $d(v) = |N(v)|$. The complement graph of a graph $G=(V,E)$ is the graph $\overline{G}=(V,\overline{E})$, where $\overline{E}=\{(u,v)|u,v \in V, u \neq v$ and $(u,v) \in E\}$.

For a subset of vertices $X \subseteq V$, we use $G[X]$ to denote the induced subgraph of $G$ whose vertex set is $X$ and edge set is the subset of $E(G)$ consisting of those edges with both endpoints in $X$. We define a network separator as the set of vertices $S \subseteq V$ whose removal decomposes a given network into $T$ connected components, denoted as $\{\mathcal{C}_1 ,\mathcal{C}_2 ,\ldots, \mathcal{C}_T\}$ with $T \geqslant 2$. It can be naturally demonstrated that $G[V \setminus S] = \cup ^T_{i=1} \mathcal{C}_i$, where $\mathcal{C}_i \cap \mathcal{C}_j = \emptyset$, $\forall i,j \in [1,2,\ldots,T]$ and $i \neq j$.

This work studies an $\alpha$-separator problem ($\alpha$-SP), which aims to find a minimum node separator (i.e., a set of vertices) whose removal divides graph $G$ into multiple different connected components, each of which has at most $\lceil \alpha \cdot n \rceil$ vertices. Formally, $\alpha$-SP can be formulated as:
\begin{alignat}{2}\label{Equ:ASP Objective Function}
    \min_{S \subseteq V}\quad& f(S)=|S| \\
    \mbox{s.t.}\quad         & G[V \setminus S] = \cup ^T_{i=1} \mathcal{C}_i\label{Equ:Component Constraint}\\
    & |\mathcal{C}_i| \leq \lceil \alpha \cdot n \rceil; \quad i = 1,\ldots,T\label{Equ:Size Constraint}
\end{alignat}
where constraint (\ref{Equ:Component Constraint}) means that the residual graph $G[V \setminus S]$ consists of $T$ connected components and the set of constraints (\ref{Equ:Size Constraint}) guarantees that each connected component $\mathcal{C}_i$ has at most $\lceil \alpha \cdot n \rceil$ nodes.

\begin{figure}[!htbp]
    \centering
    \includegraphics[width=1.0\columnwidth]{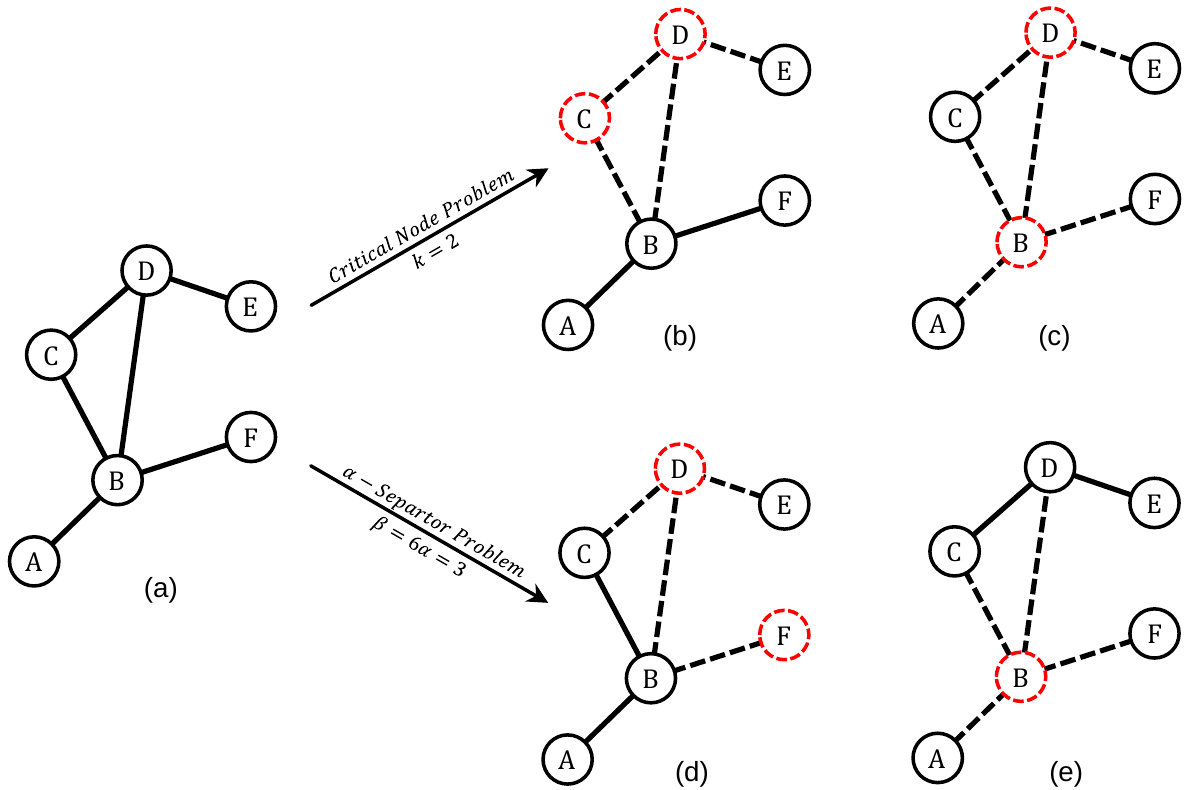}
    \caption{Illustration of the Critical Node Problem (CNP) with $k=2$ and $\alpha$-Separator Problem ($\alpha$-SP) with $\alpha=0.5$ (or $\beta=\lceil \alpha \cdot n\rceil =3$): (a) a graph with six nodes and six edges, (b) a feasible solution of CNP (i.e., $S_1=\{C,D\}$) with $\hat{f}(S_1)=3$, (c) a better solution of CNP (i.e., $S_2=\{D,B\}$) with $\hat{f}(S_2)=0$, (d) a feasible solution of $\alpha$-SP (i.e., $S_3=\{D,F\}$) with $f(S_3)=2$, (e) a better solution of $\alpha$-SP (i.e., $S_4=\{B\}$) with $f(S_4)=1$}
    \label{Fig:Problem Example}
\end{figure}

Both $\alpha$-SP and critical node problem (CNP) are two representative CNDPs, which belong to $k$-vertex-CNDP and $\beta$-connectivity-CNDP, respectively. To show their difference, we consider the following example. Figure~\ref{Fig:Problem Example}(a) shows a graph with six nodes and six edges. Assume that $k=2$ in CNP. Then the number of removed nodes should not exceed 2. Figure~\ref{Fig:Problem Example}(b) presents a feasible solution $S_1$ with $\hat{f}(S_1)=3$. Here, $\hat{f}=\sigma$ counts the number of pairs of nodes connected by a path in the residual graph. Figure~\ref{Fig:Problem Example}(c) presents a better solution $S_2$ with smaller objective function value (i.e., $\hat{f}(S_2)=0$). Suppose that $\alpha=0.5$ in $\alpha$-SP. Then $\sigma$ denotes the size of each resulting connected component that must not exceed $\lceil \alpha \cdot n \rceil = 3$ after removing some nodes. Figures~\ref{Fig:Problem Example}(d)-(e) present two feasible solutions, i.e., $S_3$ with $f(S_3)=2$ (i.e., number of removed nodes) and $S_4$ with $f(S_4)=1$. $S_4$ is better than $S_3$.

\subsection{Related Work}
\label{SubSec:Related Work}

The computational challenge and wide range of practical applications of $\alpha$-SP have attracted increasing research attention. Several solution approaches have been proposed in recent years \cite{Shen2012,Mohamed2014,Ben-Ameur2015,Lee2018,Zhou2019,Perez-Pelo2020}. They roughly fall into two categories: exact and heuristic algorithms.

Existing exact algorithms are particularly useful for solving some special topologies. For example, Shen and Smith \cite{Shen2012} applied dynamic programming algorithms to solve networks of tree structures and series-parallel graphs. Mohamed-Sidi \cite{Mohamed2014} proposed polynomial-time algorithms for networks whose topology was a tree or a cycle. Ben-Ameur et al. \cite{Ben-Ameur2015} proved that $k$-separator problem can be solved in polynomial time for some special graphs such as bounded treewidth, interval and circular-arc graphs.

In general, there is little priori knowledge about the network topology. Therefore, heuristic algorithms represent an important alternative used to find high-quality solutions within a reasonable time, especially for those cases whose solutions exact algorithms fail to deliver. Lee et al. \cite{Lee2018} proposed a random walk algorithm based on a Metropolis chain for $\alpha$-SP. P\'{e}rez-Pel\'{o} et al. \cite{Perez-Pelo2018} developed a greedy randomized adaptive search procedure (GRASP) algorithm for it. Recently, they \cite{Perez-Pelo2020} further combined GRASP with path relinking (GRASP/PR) to obtain high-quality solutions within a reasonable computing time. Zhou et al. \cite{Zhou2019} proposed a memetic algorithm (MA) for a CNP and also adopted MA to solve a cardinality-constrained CNP. To enrich the set of solution methods for solving the computationally challenging $\alpha$-SP, we propose an FIS method for it.

\section{Frequent Itemset-driven Search}
\label{Sec:Frequent Itemset Driven Search}

In this section, we present FIS for solving $\alpha$-SP. Inspired by \cite{Galinier2006}, we adopt an iterative solution strategy by considering $\alpha$-SP from the viewpoint of constraint satisfaction and solving a series of $K$-decision $\alpha$-SPs. We start from an initial legal $K$ value and solve a $K$-decision $\alpha$-SP. Once a $K$-decision $\alpha$-SP is solved, we decrease $K$ to $K$-1 and solve a new $K$-decision $\alpha$-SP again. This process is repeated until no feasible solution of a $K$-decision $\alpha$-SP can be found.

\subsection{Solution Representation and Evaluation}
\label{SubSec:Solution Representation and Evaluation}

Given an $\alpha$-SP instance with a fixed integer $K$, a candidate solution of a $K$-decision $\alpha$-SP can be represented as any subset $S \subset V$ of size $K$ (i.e., $|S|=K$). The $K$-decision $\alpha$-SP aims to find a feasible solution, i.e., a set of $K$ nodes whose removal decomposes the original $G$ into many connected components, each of which has at most $\lceil\alpha \cdot n \rceil$ nodes. $\alpha$-SP tries to minimize such $K$ value. To evaluate a candidate solution of $K$-decision $\alpha$-SP, we define a new auxiliary function $f'(S)$:
\begin{equation} \label{Equ:Auxiliary Function}
    f'(S)= \max_{\forall i \in \{1,2,\ldots,T\}}(|\mathcal{C}_i|-\lceil \alpha \cdot n \rceil,0)
\end{equation}
which only counts the nodes in the largest connected component that has over $\lceil \alpha \cdot n \rceil$ nodes. A solution $S$ is feasible if $f'(S)=0$.

\subsection{Algorithmic Framework}
\label{SubSec:Algorithmic Framework}

FIS is realized in Algorithm \ref{Alg:FIS}. It consists of five main modules: solution construction, population repair, a frequent itemset recombination (FIR), tabu search-based simulated annealing (TSSA), and rank-based population management. FIS starts from a collection of individuals with a given $K$ value, which are first built by a solution construction procedure and then modified by a population repair procedure. At each generation, an offspring solution is generated by an FIR operator and then a TSSA procedure is invoked to improve it. Once an improved offspring solution is obtained, we update the current population according to a rank-based population management strategy. If a legal solution with $K$ nodes is obtained, we reduce $K$ to $K$-1 and build a new population of individuals with new $K$ value. This process repeats until a stopping condition (e.g., reaching a time limit) is satisfied. In the following, we present each module of FIS.

\begin{algorithm}[!ht]
    \centering
    \small
    \caption{Pseudo-code of Frequent Itemset-driven Search}
    \label{Alg:FIS}
    \begin{algorithmic}[1]
        \REQUIRE{$G = (V,E)$, population size $\theta$, randomized scale factor $\eta$, and maximal iteration count $\hat{\xi}$}
        \ENSURE{The best solution found $S^*$}
        \STATE $K \leftarrow |V|$;\\
        \STATE $P \leftarrow \emptyset$;\\
        \STATE $count \leftarrow 0$;\\
        \WHILE{$count < \theta$}
            \STATE $S \leftarrow \textit{ConstructSolution}(\eta)$;\\
            \IF{$|S| < K$}
                \STATE $S^* \leftarrow S$;\\
            \ENDIF
            \STATE $P \leftarrow P \cup \{S\}$;\\
            \STATE $count \leftarrow count+1$;\\
        \ENDWHILE
        \STATE $K \leftarrow K-1$;\\
        \STATE $P \leftarrow \textit{RepairPopulation}(P,K)$;\\
        \WHILE{Stopping condition is not met}
        \STATE $S \leftarrow \textit{FIR}(P)$;\\
        \STATE $S \leftarrow \textit{TSSA}(S,\hat{\xi})$;\\
        \STATE $P \leftarrow \textit{ManagePopulation}(P,S)$;\\
        \IF{(isFeasible($S$)=True)}
        \STATE $S^* \leftarrow S$;\\
        \STATE $K \leftarrow K-1$;\\
        \STATE $P \leftarrow \textit{RepairPopulation}(P,K)$;\\
        \ENDIF
        \ENDWHILE
        \RETURN The best solution found $S^*$;
    \end{algorithmic}
\end{algorithm}

\subsection{Solution Construction}
\label{SubSec:Solution Construction}

To obtain a high-quality feasible solution, we adopt a two-stage solution construction method, as shown in Algorithm \ref{Alg:Solution Construction}. Specifically, starting from an empty set, we construct it in two stages. At the first stage, we try to obtain a feasible solution (lines 2-8). Starting from a random node (lines 2-3). Then, a node is iteratively and greedily added into the solution until a feasible solution is found (lines 4-8). At the second stage, we resort to a TSSA procedure to find a feasible solution with smaller size). We first greedily remove a node (lines 10-11), and then use TSSA to improve it (line 12). Once a feasible solution is found, we update the solution and repeat the process again; otherwise if we cannot find a feasible solution, we terminate it.

\begin{algorithm}[!ht]
    \centering
    \small
    \caption{Pseudo-code of Solution Construction Procedure}
    \label{Alg:Solution Construction}
    \begin{algorithmic}[1]
        \REQUIRE{$G = (V,E)$ and greedy scale factor $\beta$}
        \ENSURE{A feasible solution $S$}
        \STATE $S \leftarrow \emptyset$;
        \\/* Initialize a feasible solution $S$ */\\
        \STATE $u \leftarrow \textit{RandomSelect}(V,1)$;\\
        \STATE $S \leftarrow S \cup \{u\}$, $L \leftarrow V \setminus \{u\}$;\\
        \WHILE{(isFeasible($S$)=False)}
        \STATE $\mathcal{L} \leftarrow \textit{RandomSelect}(L,\eta \times |L|)$;\\
        \STATE $u \leftarrow \arg\max_{v \in \mathcal{L}}\Phi(v)$;\\
        \STATE $S \leftarrow S \cup \{u\}$, $L \leftarrow L \setminus \{u\}$;\\
        \ENDWHILE
        \\/* Find a better feasible solution $S$ of smaller $K$ value */\\
        \WHILE{True}
        \STATE $u \leftarrow \arg\min_{v \in S}f(S \setminus \{v\})$;\\
        \STATE $S' \leftarrow S \setminus \{u\}$;\\
        \STATE $S' \leftarrow \textit{TSSA}(S',\hat{\xi})$;\\
        \IF{(isFeasible($S'$)=True)}
        \STATE $S \leftarrow S'$;\\
        \ELSE
        \STATE \textbf{break};\\
        \ENDIF
        \ENDWHILE
        \RETURN A feasible solution $S$;
    \end{algorithmic}
\end{algorithm}

We define two functions in Algorithm \ref{Alg:Solution Construction}, namely isFeasible$()$ and $\Phi()$. The former is a feasible function used to check whether the legitimacy constraint of a solution is satisfied or not (lines 4 and 13), which can be finished in $O(m)$. The size of all connected components in the residual network should be no more than $\lceil \alpha \cdot n \rceil$. The latter is a greedy function, which decides the next node $u$ to be added into the partial solution $S_i$. Here, we use betweenness centrality \cite{Freeman1977} to evaluate the importance of a node to the whole network. Given a network $G=(V,E)$ and a node $u \in V$, $u$'s betweenness centrality value is calculated based on the number of shortest paths that pass through $u$, i.e.,
\begin{equation}\label{Equ:Greedy function of nodes}
    \Phi(u) = \sum_{s\ne u\ne t\in V} \frac{\sigma_{st}(u)}{\sigma_{st}}
\end{equation}
where $\sigma_{st}$ is the number of shortest path from $s$ to $t$ in $G$, and $\sigma_{st}(u)$ is the number of shortest paths from $s$ to $t$ via node $u$. Note that $\sigma_{st}=0$ means that there is no path between $s$ and $t$, and we then define $\Phi(u)=0$. In general, nodes that more frequently lie on shortest paths between other nodes have higher betweenness centrality values.

In Algorithm \ref{Alg:Solution Construction}, $\textit{RandomSelect}(para1,para2)$ is a random selection function that aims to randomly select $para2$ elements from $para1$, and $\eta \in (0,1]$ is a randomized scale factor used to control the randomness/greediness of the generated feasible solution. Specifically, $\eta \rightarrow 0$ means that only a few of nodes are selected and added into the restricted list $\mathcal{L}$, resulting in a totally random procedure. If $\eta = 1$, all nodes are added into $\mathcal{L}$ for further greedy selection, and thus it reduces to a totally greedy procedure. The higher the value of $\eta$, the more greedy the procedure is and vice versa.

\subsection{Population Repair}
\label{SubSec:Population Repair}

By independently running the solution construction procedure (Algorithm \ref{Alg:Solution Construction}) $\theta$ times, we can obtain a set of $\theta$ feasible solutions $P=\{S_1,S_2,\ldots,S_{\theta}\}$. However, these feasible solutions may have different size. To quickly obtain an initial population, we use a population repair procedure to modify them.

Our population repair module operates as follows. For each feasible solution $S_i \in P$, if its size is larger than the given target $K$, we greedily remove node $v \in S_i$ which minimally deteriorates the objective function from $S_i$, i.e., $v\gets \arg\min_{v \in S_i}f'(S_i \setminus \{v\})$. The process repeats until it only has $K$ nodes, $|S_i|=K$. After our population repair procedure, we can obtain a population of $\theta$ infeasible solutions of only $K$ nodes, i.e., $\forall S_i \in P$, $|S_i|=K$

\subsection{Frequent Itemset Recombination}
\label{SubSec:Frequent Itemset Recombination}

Frequent itemsets are a form of frequent patterns \cite{Agrawal1993} in data mining. Given transactions that are sets of items and a minimum frequency, any set of items that occurs at least in the minimum number of transactions is a frequent itemset, as shown in next.

We assume that the set of all items $\mathcal{T}=\{a,b,c,d\}$, data $\mathcal{D}=\{\{a,b,c\},\{a,c,d\},\{b,c,d\},\{b,d\},\{c,d\}\}$, and the frequency threshold $\varphi$ is 3. All possible itemsets and their frequencies are listed in Table \ref{Tab:All Possible Itemsets and Their Frequencies}. According to the given frequency threshold $\varphi=3$, then the frequent itemsets are $\{\{b\},\{c\},\{d\},\{c,d\}\}$.

\begin{table}[!ht]
\centering
\small
\caption{All Possible Itemsets and Their Frequencies of the Example}
\label{Tab:All Possible Itemsets and Their Frequencies}
\begin{threeparttable}
\setlength{\tabcolsep}{1.5mm}{
\begin{tabular}{lclclc}
\toprule[0.75pt]
Itemset      &Frequency  &Itemset     &Frequency  &Itemset     &Frequency\\
\cmidrule[0.5pt](lr){1-2} \cmidrule[0.5pt](lr){3-4} \cmidrule[0.5pt]{5-6}
$\{a\}$   & 2 & $\{a,c\}$ & 2 & $\{a,b,c\}$   & 1\\
$\textbf{\{b\}}$   & \textbf{3} & $\{a,d\}$ & 1 & $\{a,b,d\}$   & 0\\
$\textbf{\{c\}}$   & \textbf{4} & $\{b,c\}$ & 2 & $\{a,c,d\}$   & 1\\
$\textbf{\{d\}}$   & \textbf{4} & $\{b,d\}$ & 2 & $\{b,c,d\}$   & 1\\
$\{a,b\}$ & 1 & $\textbf{\{c,d\}}$ & \textbf{3} & $\{a,b,c,d\}$ & 0\\
\bottomrule[0.75pt]
\end{tabular}}
\begin{tablenotes}
    \item Note that frequent itemsets are in bold.
\end{tablenotes}
\end{threeparttable}
\end{table}

Our $\alpha$-SP is a typical subset selection problem \cite{Qian2019,Zhu2020} that concerns finding a subset of a given set such that a given set of constraints is satisfied. It is well-known that some common elements appear in many high-quality solutions during a search. Among high-quality solutions, they are closely related to frequent itemsets. Inspired by the use of frequent itemsets in data mining \cite{Luna2019}, we propose a frequent itemset recombination operator to generate offspring based on the common elements shared by high-quality solutions during a search. We treat high-quality solutions as transactions, and each transaction has same size. The set of common elements shared by high-quality solutions can be naturally modelled as a frequent itemset. Then, an offspring solution can be generated based on a mined frequent itemset. Algorithm \ref{Alg:FIR} demonstrates the pseudo code of FIR operator. It has four main steps:
\begin{enumerate}
    \item Build a reference solution set $\overline{P}$ by randomly selecting $\overline{\theta}$ solutions from $P$, and count the frequency of each node. Formally, we calculate the frequency of each node $v$ as follows:
        \begin{equation}\label{Equ:Frequency of Nodes}
            \Psi(v,\overline{P}) =\sum_{S_i \in \overline{P}}\chi(v,S_i)
        \end{equation}
        where $\chi(v,S_i)=1$ if $v \in S_i$, and otherwise 0.
    \item Choose a solution from $\hat{P}$ as a base solution $S_b$ in a random way, where $\hat{P}$ is the elite solution set that consists of $\hat{\theta}$ highest-quality solutions selected from $P$;
    \item Construct a partial solution $S_o$ by directly inheriting an frequent itemset (i.e., $\lfloor \rho \times K \rfloor$ nodes that appear most frequently in $\overline{P}$), where $\rho \in (0,1]$ is a frequent itemset scale factor;
    \item Repair the partial solution $S_o$ until its size $|S_o|=K$.
\end{enumerate}

\begin{algorithm}[!ht]
    \centering
    \small
    \caption{Pseudo-code of Frequent Itemset Recombination Operator}
    \label{Alg:FIR}
    \begin{algorithmic}[1]
        \REQUIRE{Population $P$ and itemset scale factor $\rho$}
        \ENSURE{An offspring solution $S_o$}
        \\/*Build a reference set $\overline{P}$*/\\
        \STATE $\overline{P} \leftarrow \textit{RandomSelect}(P,\overline{\theta})$;
        \\/*Choose a base solution $S_b$ from $\hat{P}$*/\\
        \STATE $S_b \leftarrow \textit{RandomSelect}(\hat{P},1)$;
        \\/*Construct a partial solution $S_o$*/\\
        \STATE $S_o \leftarrow \emptyset$;\\
        \WHILE{($|S_o|< \rho \times K$)}
        \STATE $u \leftarrow \arg\max_{v \in S_b \land v\notin S_o}\Psi(v,\overline{P})$;\\
        \STATE $S_o \leftarrow S_o \cup \{u\}$;\\
        \ENDWHILE
        \\/*Repair the partial solution $S_o$*/
        \WHILE{($|S_o| < K$)}
        \STATE $\mathcal{\hat{L}} \leftarrow \cup_{|\mathcal{C}_i|>\lceil \alpha\cdot n \rceil} \mathcal{C}_i$;\\
        \STATE $v \leftarrow \textit{RandomSelect}(\mathcal{\hat{L}},1)$;\\
        \STATE $S_o \leftarrow S_o \cup \{v\}$;\\
        \ENDWHILE
        \RETURN An offspring solution $S_o$;
    \end{algorithmic}
\end{algorithm}

Note that our frequent itemset among high-quality solutions is identified based on the frequencies of nodes in $\overline{P}$ instead of using a time-consuming data mining algorithm, e.g., FPmax* \cite{Luna2019}. To better understand the basic idea of FIR, we illustrate it with the following example.

\begin{figure}[!htbp]
    \centering
    \includegraphics[width=0.75\columnwidth]{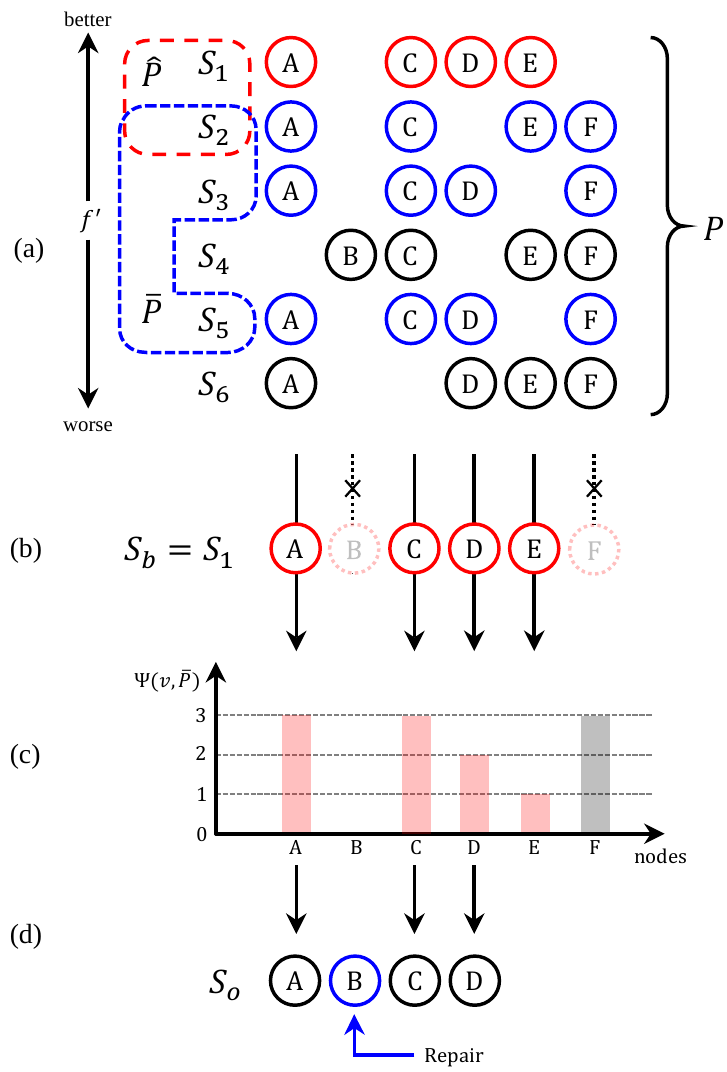}
    \caption{Diagram of FIR Operator to Generate an Offspring Solution}
    \label{Fig:An Illustrative Diagram of FIR Operator to Generate an Offspring Solution}
\end{figure}

As shown in Figure~\ref{Fig:An Illustrative Diagram of FIR Operator to Generate an Offspring Solution}, there is a population of six individuals, i.e., $P = \{S_1,S_2,S_3,S_4,S_5,S_6\}$, where $\overline{P} =\{S_2,S_3,S_5\}$ and $\hat{P}=\{S_1,S_2\}$. We first randomly select one solution from $\hat{P}$ and treat it as a base solution, i.e., $S_b = S_1$. Then, we construct a partial solution $S_o$ by directly inheriting a frequent itemset of $\lfloor 0.95 \times 4 \rfloor = 3$ nodes that appear most frequently in $\overline{P}$, i.e., $S_o=\{A,C,D\}$. Finally, we repair it in a random way until an offspring solution of $K$ nodes is obtained, i.e., $S_o = \{A,B,C,D\}$.

\subsection{Tabu Search-based Simulated Annealing}
\label{SubSec:Tabu Search-based Simulated Annealing}

To ensure an effective examination of the search space, FIS employs an effective TSSA procedure. It integrates the tabu search strategy into the classic simulated annealing (SA) framework. Algorithm \ref{Alg:TSSA} realizes it. At each iteration, a neighboring solution is first obtained based on a two-phase node exchange strategy and an attribute-based tabu search strategy. Then, a solution acceptance strategy is applied to accept or discard it. This process repeats until a feasible solution is found or a given stopping condition is satisfied.

\begin{algorithm}[!ht]
    \centering
    \small
    \caption{Pseudo-code of Tabu Search-based Simulated Annealing Module}
    \label{Alg:TSSA}
    \begin{algorithmic}[1]
        \REQUIRE{A solution $S$ and maximal iteration count $\hat{\xi}$}
        \ENSURE{The improved solution $S$}
        \STATE $\mathcal{\overline{L}} \leftarrow \emptyset$;\\
        \STATE $\xi \leftarrow 0$;\\
        \WHILE{$(\xi < \hat{\xi})$}
        \STATE $S' \leftarrow S$;
        \\/* Generate a neighboring solution*/\\
        \STATE $\mathcal{\hat{L}} \leftarrow \{u:u \notin S \land u \notin \mathcal{\overline{L}} \}$;\\
        \STATE $u \leftarrow \textit{RandomSelect}(\mathcal{\hat{L}},1)$;\\
        \STATE $S' \leftarrow S' \cup \{u\}$;\\
        \STATE $v \leftarrow \arg\min_{w \in S'} f'(S' \setminus \{w\})$;\\
        \STATE $S' \leftarrow S' \setminus \{v\}$;\\
        \IF{(isFeasible($S'$)=True)}
        \STATE $S \leftarrow S'$;\\
        \RETURN $S$;\\
        \ENDIF
        \\/* Accept or discard the solution*/\\
        \STATE $\Delta f = f'(S')-f'(S)$;\\
        \IF{($\Delta f < 0$)}
        \STATE $S \leftarrow S'$;\\
        \STATE $\xi \leftarrow 0$;\\
        \ELSE
        \IF{($rand() < p(\xi,S,S')$)}
        \STATE $S \leftarrow S'$;\\
        \STATE $\xi \leftarrow \xi + 1$;\\
        \STATE $\mathcal{\overline{L}} \leftarrow \mathcal{\overline{L}} \cup \{u\}$;\\
        \ELSE
        \STATE $\xi \leftarrow \xi + 1$;\\
        \STATE $\mathcal{\overline{L}} \leftarrow \mathcal{\overline{L}} \cup \{u\}$;\\
        \ENDIF
        \ENDIF
        \ENDWHILE
        \RETURN The improved solution $S$;
    \end{algorithmic}
\end{algorithm}

\subsubsection{Two-phase Node Exchange}
\label{SubSubSec:Two-Phase Node Exchange}

The performance of a local search procedure greatly depends on its neighborhood structure \cite{Wu2021a}. Given a candidate solution $S$ of $\alpha$-SP, a traditional neighborhood $N(S)$ can be formally defined as the set of solutions obtained from $S$ by applying operation $Swap(S,u,v) = S\cup \{u\} \setminus\{v\}$, i.e.,
\begin{equation}\label{Equ:NeighborhoodDef}
    N(S) = \{Swap(S,u,v) | \ \forall u \in V \setminus S, v \in S  \}
\end{equation}

Inspired by \cite{Zhou2019}, we also adopt a two-phase node exchange strategy. It breaks a swap operation on a node pair into two phases: an ``add phase'' aims to add a node $v$ into the current solution; and a ``remove phase'' tires to remove a node $u$ from it.
\begin{itemize}
    \item Add phase: We first construct a candidate list $\mathcal{\hat{L}} =\{v|v \notin S' \land v \notin \mathcal{\overline{L}}\}$ where $\mathcal{\overline{L}}$ is a tabu list. Then, we randomly select a node $u \in \mathcal{\overline{L}}$ and add it into the solution, i.e., $S' \leftarrow S' \cup \{u\}$.
    \item Remove phase: For each $v \in S'$, we calculate its objective value $f(S' \setminus \{v\})$. The node $v$ with the minimum objective function value is then selected and removed from the solution, i.e., $S' \leftarrow S' \setminus \{v\}$.
\end{itemize}

\subsubsection{Tabu Search Strategy}
\label{SubSubSec:Tabu Search Strategy}

To escape from local optima, an attribute-based dynamic tabu search strategy is integrated into SA. Such strategy was originally proposed in \cite{Glover1989}. Here, we use a tabu list $\mathcal{\overline{L}}$ with the dynamic tabu tenure $\gamma \cdot (n-K)$ to record the forbidden nodes, where $\gamma \in [0,1)$ is a tabu scale factor. At each iteration, if the resulting solution $S' \leftarrow S \cup \{u\} \setminus \{v\}$ is worse than $S$, $u$ is added into $\mathcal{\overline{L}}$. Once $\mathcal{\overline{L}}$ is full, the first node added into it is removed from it.

Unlike the traditional strategy that forbids all performed operations, our attribute-based strategy only forbids the operations that cause solution deterioration. Therefore, the nodes whose exchange resulting in poor solution quality has a smaller possibility to be visited again in the following iterations.

\subsubsection{Solution Acceptance Criterion}
\label{SubSubSec:Solution Acceptance Criterion}

Once a neighboring solution is obtained, we then accept or discard it according to a solution acceptance criterion. It shares a similar idea with the acceptance rule of SA \cite{Kirkpatrick1983}. The acceptance of each candidate $S'$ is subject to the probability $p(\xi,S,S')$ defined as follows:
\begin{equation} \label{Equ:Accetpance Rule}
p(\xi,S,S') =
    \begin{cases}
    {1}                                                   & f(S') < f(S)       \\
    {e^{-(f(S')-f(S)) \cdot \frac{\xi}{\hat{\xi}}}}, & \textrm{otherwise}
    \end{cases}
\end{equation}
where $\hat{\xi}$ and $\xi$ represent the maximal iteration count and the current iteration count, respectively. Note that our acceptance criterion is different from the Metropolis condition of SA, especially for determining the probability of accepting a worse solution.

\subsection{Rank-based Population Management}
\label{SubSec:Rank-based Population Management}

Inspired by \cite{Fu2015,Zhou2021a}, we resort to a quality-and-distance population management strategy to maintain a healthy population. It simultaneously considers the solution quality and solution distance during an evolutionary search. We define a combined measure between solution quality and solution distance as follows:
\begin{equation}\label{Equ:Score Function}
    Q(S_i,P') = \mu\cdot R_{f'}(S_i)+(1-\mu)\cdot R_d(S_i)
\end{equation}
where $S_i$ is the $i$-th solution, $\mu$ is a weight factor, $R_{f'}()$ and $R_d()$ are two functions used to calculate the ranks of a given solution in the extended population $P'$ according to the auxiliary objective function $f'()$ and distance function $d()$, respectively. Here, we use the Manhattan distance as the solution distance metric, and $d()$ is formally defined as the total sum of the distances between $S_i$ and other solutions $S_j$ in $P'$, i.e.,
\begin{equation} \label{Equ:Distance Metric}
    d(S_i)=\sum_{S_j \in P' \setminus \{S_i\}} (|S_i \cup S_j|-|S_i \cap S_j|)
\end{equation}

Given a solution $S$ and the current population $P=\{S_1,S_2,\ldots,S_{\theta}\}$, our rank-based population management strategy aims to decide whether $S$ should be inserted into $P$ or discarded, which operates as follows: 1) We temporarily add $S$ into population $P$, thus obtaining an extended population $P'$, i.e., $P' \leftarrow P \cup \{S\}$; 2) We evaluate each individual of population $P'$ according to the following score function (see Equation (\ref{Equ:Score Function})) and find out the worst solution $S_w$, i.e., $S_w \leftarrow \arg\max_{S_i \in P'}Q(S_i,P')$; and 3) We compare $S_w$ with $S$. If $S_w$ is different from $S$, we replace $S_w$ by $S$. Otherwise, we discard $S$.

\subsection{Computational Complexity of FIS}
\label{SubSec:Computational Complexity of FIS}

To analyze the computational complexity of FIS, we consider its five main modules: solution construction, population repair, FIR, TSSA, and rank-based population management. Their time complexities are listed in Table~\ref{Tab:Complexity of Main Modules of FIS}.

\begin{table}[!ht]
\centering
\small
\caption{Complexity of Main Modules of FIS}
\label{Tab:Complexity of Main Modules of FIS}
\begin{threeparttable}
\setlength{\tabcolsep}{1.0mm}{
\begin{tabular}{llc}
\toprule[0.75pt]
Module      &Complexity               &Section  \\
\midrule[0.5pt]
Population Construction &$O(mn + (m+n)K)$ &\ref{SubSec:Solution Construction} \\
Population Repair &$O(m)$   &\ref{SubSec:Population Repair} \\
FIR&$O(n\log{(K)}+KT)$&\ref{SubSec:Frequent Itemset Recombination}\\
TSSA&$O((m+n) \cdot \overline{\xi})$&\ref{SubSec:Tabu Search-based Simulated Annealing}\\
Rank-based Population Management&$O(K + T)$&\ref{SubSec:Rank-based Population Management}\\
\bottomrule[0.75pt]
\end{tabular}}
\begin{tablenotes}
    \item where $\overline{\xi}$ is the total number of iterations in TSSA.
\end{tablenotes}
\end{threeparttable}
\end{table}

At the beginning of FIS, $\theta$ high-quality feasible solutions are obtained by using a solution construction procedure, following by a population repair procedure. Then, at each generation, it iteratively executes FIR, TSSA, rank-based population management and population repair until a given stopping condition is satisfied. The total complexity of each generation is $O(n\log{(K)}+KT+(m+n) \cdot \overline{\xi})$.

\section{Empirical Results}
\label{Sec:Empirical Results}

We conduct extensive experiments to evaluate FIS. Our targets are to 1) demonstrate the benefit of TSSA over the best-performing local search algorithm \cite{Zhou2019} and 2) evaluate the performance of FIS with respect to state-of-the-art algorithms \cite{Perez-Pelo2018,Zhou2019,Perez-Pelo2020} on benchmark instances.

\subsection{Experimental Settings and Benchmarks}
\label{SubSec:Experimental Settings and Benchmarks}

\subsubsection{Parameter Settings}
\label{SubSubSec:Parameter Settings}

We carry out extensive experiments to evaluate the performance of FIS\footnote{Our programs and results will be made available at \url{https://github.com/YangmingZhou/AlphaSeparatorProblem} once this paper is accepted.}. All algorithms are implemented in C++ and complied by g++ with `-Ofast -march=native'. They are run on a server with Intel E5-2680 v2 \@ 2.8 GHz CPU ($\times 2$) and 64 GB RAM under Windows 10 Enterprise OS. Table \ref{Tab:Parameter Settings} gives the main parameter settings of FIS. They can be divided into two categories according to our preliminary experimental results. The first category includes population size $\theta$, reference solution set size $\overline{\theta}$ and elite solution set size $\hat{\theta}$, which are determined based on our preliminary experimental results. The second category includes randomized scale factor $\eta$, frequent itemset scale factor $\rho$, maximal iteration count $\hat{\xi}$, tabu scale factor $\gamma$ and weight factor $\mu$. They are more sensitive than those in the first category. To tune them, we resort to the well-known automatic parameter configuration tool, IRACE \cite{Lopez2016}. Section \ref{SubSec:Parameter Sensitivity Analysis} gives the detailed parameter sensitivity analysis. Note that we solve ER\_100 and ER\_200 instances with time limits $\hat{t}=30$ seconds and $\hat{t}=500$ seconds, respectively.

\begin{table}[!ht]
\centering
\small
\caption{Parameter Settings of FIS}
\label{Tab:Parameter Settings}
\setlength{\tabcolsep}{1.5mm}{
\begin{tabular}{lllc}
\toprule[0.75pt]
Parameter  & Description             & Final Value & Section \\
\midrule[0.5pt]
$\theta$   & Population Size         & 50          &\ref{SubSec:Algorithmic Framework}                 \\
$\overline{\theta}$ & Reference Set Size  & 0.5$\theta$ &\ref{SubSec:Frequent Itemset Recombination} \\
$\hat{\theta}$ & Elite Set Size      & 0.1$\theta$ &\ref{SubSec:Frequent Itemset Recombination} \\
$\eta$     & Randomized Scale Factor & 0.6         &\ref{SubSec:Solution Construction}                 \\
$\rho$     & Itemset Scale Factor    & 0.95        &\ref{SubSec:Frequent Itemset Recombination} \\
$\hat{\xi}$& Maximal Iteration Count & 2000        &\ref{SubSec:Tabu Search-based Simulated Annealing} \\
$\gamma$   & Tabu Scale Factor       & 0.2         &\ref{SubSec:Tabu Search-based Simulated Annealing} \\
$\mu$      & Weight Factor           & 0.6         &\ref{SubSec:Rank-based Population Management} \\
        \bottomrule[0.75pt]
\end{tabular}}
\end{table}

\subsubsection{Benchmarks}
\label{SubSubSec:Benchmarks}

Following the studies \cite{Perez-Pelo2018,Zhou2019,Perez-Pelo2020}, our experiments are conducted on 50 widely used benchmark instances. These instances fall into two categories: ER\_100 and ER\_200. They were originally presented in \cite{Perez-Pelo2018}. Each graph is generated by using the Erd\"{o}s R\'{e}nyi model \cite{Erdos2011}, and each new inserted vertex has the same probability of being connected to any existent vertex in the graph. Since $\alpha$-SP is NP-hard for general topology networks only when $\alpha \leq 2/3$ \cite{Feige2006}. For each instance, we solve the $\alpha$-SP for values of $\alpha \in \{0.2,0.4,0.6\}$ respectively in the following experiments.

\subsubsection{Statistical Tests}
\label{SubSubSec:Statistical Tests}

In this work, we use two kinds of statistical tests recommended by Dem\v{s} \cite{Demsar2006}. We resort to the well-known \emph{Wilcoxon signed ranks test} to check the significant difference on each comparison indicator between two algorithms. Given two algorithms: A and B, at a significance level of 0.05, there is a significant performance difference between algorithm A and algorithm B if the computed p-value is less than 0.05. We use a two-step statistic test procedure for comparisons of multiple algorithms. We first conduct a \emph{Friedman test} which makes the null hypothesis that all algorithms are equivalent. If such hypothesis is rejected, we then proceed with the \emph{two-tailed Bonferroni-Dunn test} to check the significant difference among these algorithms.

\subsection{Comparative Results between TSSA and CBNS}
\label{SubSec:Comparative Results Between TSSA and CTAS}

To show the benefit of TSSA, we experimentally compare it with a powerful local search named component-based neighborhood search (CBNS). CBNS was originally proposed in \cite{Zhou2019}. It effectively integrates a component-based node exchange strategy and a node weighting scheme, and demonstrates excellent performance for solving critical node problems. To make a fair comparison between them, we run both algorithms under the same conditions in our computational platform. The comparative results on 28 ER\_100 instances and 22 ER\_200 instances are reported in Tables \ref{Tab:Comparations Between TSSA and CBNS on 28 ER100 Instances} and \ref{Tab:Comparations Between TSSA and CBNS on 22 ER200 Instances}, respectively.

\begin{table}[!hbtp]
    \centering
    \small
    \caption{Comparison between TSSA and CBNS on 28 ER\_100 Instances}
    \label{Tab:Comparations Between TSSA and CBNS on 28 ER100 Instances}
    \begin{threeparttable}
    \setlength{\tabcolsep}{1.0mm}{
        \begin{tabular}{lc|rrr|rrr}
            \toprule[0.75pt]
            \multicolumn{2}{c}{}         & \multicolumn{3}{c}{CBNS}     & \multicolumn{3}{c}{\textbf{TSSA}}                                                                                                                                                                                     \\
            \cmidrule[0.5pt](lr){3-5} \cmidrule[0.5pt](lr){6-8}
            \multicolumn{1}{l}{Instance} & \multicolumn{1}{c}{$\alpha$} & \multicolumn{1}{c}{$\hat{f}$}     & \multicolumn{1}{c}{$\overline{f}$} & \multicolumn{1}{c}{$\overline{t}$} & \multicolumn{1}{c}{$\hat{f}$} & \multicolumn{1}{c}{$\overline{f}$} & \multicolumn{1}{c}{$\overline{t}$} \\
            \midrule[0.5pt]
            ER\_100\_0.05\_0.2\_0        & 0.2                          & \textbf{26}                       & 28.27                              & 0.06                               & \textbf{26}                   & \textbf{26.00}                     & 0.24                               \\
            ER\_100\_0.05\_0.2\_2        & 0.2                          & \textbf{27}                       & 28.70                              & 0.04                               & \textbf{27}                   & \textbf{27.00}                     & 0.47                               \\
            ER\_100\_0.05\_0.2\_3        & 0.2                          & 32                                & 33.17                              & 0.01                               & \textbf{31}                   & \textbf{31.00}                     & 0.31                               \\
            ER\_100\_0.05\_0.2\_4        & 0.2                          & 29                                & 30.00                              & 0.01                               & \textbf{28}                   & \textbf{28.00}                     & 6.81                               \\
            ER\_100\_0.05\_0.5\_0        & 0.4                          & \textbf{25}                       & 27.20                              & 0.01                               & \textbf{25}                   & \textbf{25.93}                     & 0.69                               \\
            ER\_100\_0.05\_0.5\_1        & 0.4                          & \textbf{22}                       & 24.80                              & 0.01                               & \textbf{22}                   & \textbf{22.00}                     & 0.12                               \\
            ER\_100\_0.05\_0.5\_3        & 0.4                          & \textbf{22}                       & 24.17                              & 0.01                               & \textbf{22}                   & \textbf{22.00}                     & 0.31                               \\
            ER\_100\_0.05\_0.8\_0        & 0.6                          & \textbf{19}                       & 21.23                              & 0.01                               & \textbf{19}                   & \textbf{19.00}                     & 0.40                               \\
            ER\_100\_0.05\_0.8\_3        & 0.6                          & \textbf{17}                       & 19.20                              & 0.01                               & \textbf{17}                   & \textbf{17.00}                     & 0.17                               \\
            ER\_100\_0.06\_0.2\_3        & 0.2                          & 36                                & 37.67                              & 0.01                               & \textbf{35}                   & \textbf{35.63}                     & 5.68                               \\
            ER\_100\_0.06\_0.2\_4        & 0.2                          & 36                                & 37.43                              & 0.01                               & \textbf{35}                   & \textbf{35.63}                     & 3.62                               \\
            ER\_100\_0.06\_0.5\_3        & 0.4                          & 30                                & 32.27                              & 0.01                               & \textbf{29}                   & \textbf{29.00}                     & 1.63                               \\
            ER\_100\_0.06\_0.8\_4        & 0.6                          & \textbf{20}                       & 22.07                              & 0.01                               & \textbf{20}                   & \textbf{20.00}                     & 0.64                               \\
            ER\_100\_0.07\_0.2\_1        & 0.2                          & \textbf{39}                       & 41.40                              & 0.01                               & \textbf{39}                   & \textbf{39.00}                     & 2.86                               \\
            ER\_100\_0.07\_0.2\_4        & 0.2                          & \textbf{36}                       & 38.20                              & 0.01                               & \textbf{36}                   & \textbf{36.00}                     & 9.26                               \\
            ER\_100\_0.07\_0.5\_0        & 0.4                          & \textbf{34}                       & 36.37                              & 0.01                               & \textbf{34}                   & \textbf{34.00}                     & 0.85                               \\
            ER\_100\_0.07\_0.5\_2        & 0.4                          & 32                                & 34.73                              & 0.01                               & \textbf{31}                   & \textbf{31.00}                     & 2.12                               \\
            ER\_100\_0.07\_0.8\_0        & 0.6                          & \textbf{24}                       & 26.50                              & 0.02                               & \textbf{24}                   & \textbf{24.00}                     & 2.27                               \\
            ER\_100\_0.07\_0.8\_4        & 0.6                          & \textbf{25}                       & 26.90                              & 0.02                               & \textbf{25}                   & \textbf{25.00}                     & 1.80                               \\
            ER\_100\_0.08\_0.5\_3        & 0.4                          & 39                                & 40.80                              & 0.01                               & \textbf{38}                   & \textbf{38.00}                     & 2.24                               \\
            ER\_100\_0.08\_0.5\_4        & 0.4                          & 37                                & 39.27                              & 0.01                               & \textbf{36}                   & \textbf{36.00}                     & 0.91                               \\
            ER\_100\_0.08\_0.8\_2        & 0.6                          & \textbf{27}                       & 28.93                              & 0.04                               & \textbf{27}                   & \textbf{27.00}                     & 1.29                               \\
            ER\_100\_0.08\_0.8\_4        & 0.6                          & \textbf{28}                       & 29.87                              & 0.08                               & \textbf{28}                   & \textbf{28.00}                     & 3.62                               \\
            ER\_100\_0.09\_0.2\_1        & 0.2                          & \textbf{45}                       & 47.43                              & 0.01                               & \textbf{45}                   & \textbf{45.07}                     & 7.47                               \\
            ER\_100\_0.09\_0.2\_4        & 0.2                          & \textbf{45}                       & 47.43                              & 0.01                               & \textbf{45}                   & \textbf{45.00}                     & 2.75                               \\
            ER\_100\_0.09\_0.5\_2        & 0.4                          & 40                                & 42.40                              & 0.01                               & \textbf{39}                   & \textbf{39.00}                     & 4.99                               \\
            ER\_100\_0.09\_0.8\_1        & 0.6                          & 28                                & 28.47                              & 0.06                               & \textbf{27}                   & \textbf{27.17}                     & 8.18                               \\
            ER\_100\_0.09\_0.8\_2        & 0.6                          & \textbf{26}                       & 27.03                              & 0.06                               & \textbf{26}                   & \textbf{26.00}                     & 0.19                               \\
            \midrule[0.5pt]
            \#Wins                       & $-$                          & 0                                 & 0                                  & $-$                                & 10                            & 28                                 & $-$                                \\
            \#Ties                       & $-$                          & 18                                & 0                                  & $-$                                & 28                            & 0                                  & $-$                                \\
            \#Loses                      & $-$                          & 10                                & 28                                 & $-$                                & 0                             & 0                                  & $-$                                \\
            \midrule[0.5pt]
            p-value                      & $-$                          & \textbf{7.8E-4}                   & \textbf{1.9E-6}                    & $-$                                & $-$                           & $-$                                & $-$                                \\
            \bottomrule[0.75pt]
        \end{tabular}}
    \end{threeparttable}
\end{table}

\begin{table}[!hbtp]
    \centering
    \small
    \caption{Comparison between TSSA and CBNS on 22 ER\_200 Instances}
    \label{Tab:Comparations Between TSSA and CBNS on 22 ER200 Instances}
    \begin{threeparttable}
    \setlength{\tabcolsep}{1.0mm}{
        \begin{tabular}{lc|rrr|rrr}
            \toprule[0.75pt]
            \multicolumn{2}{c}{}         & \multicolumn{3}{c}{CBNS}     & \multicolumn{3}{c}{\textbf{TSSA}}                                                                                                                                                                                     \\
            \cmidrule[0.5pt](lr){3-5} \cmidrule[0.5pt](lr){6-8}
            \multicolumn{1}{l}{Instance} & \multicolumn{1}{c}{$\alpha$} & \multicolumn{1}{c}{$\hat{f}$}     & \multicolumn{1}{c}{$\overline{f}$} & \multicolumn{1}{c}{$\overline{t}$} & \multicolumn{1}{c}{$\hat{f}$} & \multicolumn{1}{c}{$\overline{f}$} & \multicolumn{1}{c}{$\overline{t}$} \\
            \midrule[0.5pt]
            ER\_200\_0.05\_0.2\_0        & 0.2                          & 95                                & 97.73                              & 0.04                               & \textbf{92}                   & \textbf{93.03}                     & 112.32                             \\
            ER\_200\_0.05\_0.2\_1        & 0.2                          & 95                                & 97.23                              & 0.04                               & \textbf{93}                   & \textbf{93.63}                     & 116.08                             \\
            ER\_200\_0.05\_0.2\_3        & 0.2                          & 94                                & 97.03                              & 0.04                               & \textbf{93}                   & \textbf{93.17}                     & 205.47                             \\
            ER\_200\_0.05\_0.5\_0        & 0.4                          & 86                                & 89.63                              & 0.03                               & \textbf{83}                   & \textbf{83.00}                     & 185.43                             \\
            ER\_200\_0.05\_0.5\_4        & 0.4                          & 83                                & 85.27                              & 0.03                               & \textbf{78}                   & \textbf{78.97}                     & 211.54                             \\
            ER\_200\_0.05\_0.8\_3        & 0.6                          & 60                                & 61.40                              & 1.76                               & \textbf{59}                   & \textbf{59.00}                     & 72.20                              \\
            ER\_200\_0.06\_0.5\_1        & 0.4                          & 92                                & 93.97                              & 0.09                               & \textbf{88}                   & \textbf{89.03}                     & 185.77                             \\
            ER\_200\_0.06\_0.5\_2        & 0.4                          & 94                                & 96.47                              & 0.14                               & \textbf{91}                   & \textbf{91.53}                     & 125.33                             \\
            ER\_200\_0.06\_0.8\_1        & 0.6                          & 64                                & 65.17                              & 3.63                               & \textbf{63}                   & \textbf{63.93}                     & 174.27                             \\
            ER\_200\_0.06\_0.8\_4        & 0.6                          & \textbf{64}                       & 65.33                              & 7.10                               & 65                            & \textbf{65.30}                     & 184.37                             \\
            ER\_200\_0.07\_0.2\_1        & 0.2                          & 110                               & 112.10                             & 0.11                               & \textbf{109}                  & \textbf{109.43}                    & 136.15                             \\
            ER\_200\_0.07\_0.5\_1        & 0.4                          & 96                                & 97.97                              & 0.33                               & \textbf{93}                   & \textbf{93.00}                     & 105.67                             \\
            ER\_200\_0.07\_0.5\_4        & 0.4                          & 95                                & 96.93                              & 0.38                               & \textbf{93}                   & \textbf{93.03}                     & 170.44                             \\
            ER\_200\_0.07\_0.8\_3        & 0.6                          & \textbf{66}                       & \textbf{67.80}                     & 6.71                               & 69                            & 69.27                              & 223.27                             \\
            ER\_200\_0.08\_0.2\_4        & 0.2                          & \textbf{115}                      & 120.37                             & 0.07                               & \textbf{115}                  & \textbf{115.37}                    & 144.68                             \\
            ER\_200\_0.08\_0.5\_0        & 0.4                          & 100                               & 102.43                             & 1.24                               & \textbf{98}                   & \textbf{99.00}                     & 201.08                             \\
            ER\_200\_0.08\_0.5\_2        & 0.4                          & 98                                & 100.40                             & 0.41                               & \textbf{97}                   & \textbf{97.00}                     & 94.60                              \\
            ER\_200\_0.08\_0.8\_3        & 0.6                          & \textbf{68}                       & \textbf{68.97}                     & 10.33                              & 71                            & 71.93                              & 255.68                             \\
            ER\_200\_0.08\_0.8\_4        & 0.6                          & 68                                & \textbf{69.83}                     & 9.76                               & 73                            & 73.43                              & 182.74                             \\
            ER\_200\_0.09\_0.2\_1        & 0.2                          & 122                               & 126.67                             & 0.05                               & \textbf{119}                  & \textbf{120.30}                    & 186.39                             \\
            ER\_200\_0.09\_0.5\_1        & 0.4                          & \textbf{101}                      & 103.27                             & 1.32                               & \textbf{101}                  & \textbf{101.47}                    & 150.14                             \\
            ER\_200\_0.09\_0.8\_3        & 0.6                          & \textbf{70}                       & \textbf{71.03}                     & 14.08                              & 74                            & 74.60                              & 104.88                             \\
            \midrule[0.5pt]
            \#Wins                       & $-$                          & 5                                 & 4                                  & $-$                                & 15                            & 18                                 & $-$                                \\
            \#Ties                       & $-$                          & 2                                 & 0                                  & $-$                                & 2                             & 0                                  & $-$                                \\
            \#Loses                      & $-$                          & 15                                & 18                                 & $-$                                & 5                             & 4                                  & $-$                                \\
            \midrule[0.5pt]
            p-value                      & $-$                          & 7.4E-2                            & \textbf{6.3E-4}                    & $-$                                & $-$                           & $-$                                & $-$                                \\
            \bottomrule[0.75pt]
        \end{tabular}}
    \end{threeparttable}
\end{table}

In Tables \ref{Tab:Comparations Between TSSA and CBNS on 28 ER100 Instances} and \ref{Tab:Comparations Between TSSA and CBNS on 22 ER200 Instances}, columns 1 and 2 present for each instance its name (Instance) and corresponding $\alpha$ value, respectively. Columns 3-5 report the results of CBNS, i.e., the best result ($\hat{f}$) found during 30 runs, average result ($\overline{f}$), and average time ($\overline{t}$) in seconds required to research the best result at each run. Similarly, columns 6-8 give the result of TSSA. The better value of each pair of compared results is indicated in bold. In addition, we give the number of instances on which one algorithm's result is a better than (\#Wins), equal to (\#Ties), and worse than (\#Loses) another's. At its bottom, we also provide the p-values of the Wilcoxon signed ranks test.

From Table \ref{Tab:Comparations Between TSSA and CBNS on 28 ER100 Instances}, we observe that TSSA demonstrates excellent performance on 28 instances of ER\_100. Compared to CBNS, it find better results on 8 out of 28 instances, and reaches the equal results on remaining 20 instances in terms of $\hat{f}$. In terms of $\overline{f}$, TSSA outperforms CBNS on all 28 instances. At a significance level of 0.05, it significantly better than CBNS in terms of both $\hat{f}$ and $\overline{f}$.

TSSA also demonstrates excellent performance on 22 ER\_200 instances, as shown in Table \ref{Tab:Comparations Between TSSA and CBNS on 22 ER200 Instances}. In particular, it finds better $\hat{f}$ values on 13 out of 22 instances, matches equal $\hat{f}$ values on 6 instances, and worse $\hat{f}$ values on 3 remaining instances. While for the performance indicator $\overline{f}$, it obtains better values on 14 out of 22 instances, and worse values on remaining 8 instances. At a significance level of 0.05, TSSA significantly outperforms CBNS in terms of $\overline{f}$. These interesting observations from Tables \ref{Tab:Comparations Between TSSA and CBNS on 28 ER100 Instances} and \ref{Tab:Comparations Between TSSA and CBNS on 22 ER200 Instances} confirm TSSA's superiority over CBNS.

\subsection{Comparative Results between FIS and State-of-the-art Algorithms}
\label{SubSec:Comparative Results Among FIS and State-of-the-art Algorithms}

To evaluate the performance of FIS, we carry out detailed comparisons between FIS and three recent state-of-the-art algorithms: greedy randomized adaptive search procedure (GRASP) \cite{Perez-Pelo2018}, memetic algorithm (MA) \cite{Zhou2019}, and GRASP with path relinking (GRASP/PR) \cite{Perez-Pelo2020}.

Since the source codes of GRASP and GRASP/PR are not available to us, we re-implement them according to \cite{Perez-Pelo2018} and \cite{Perez-Pelo2020}. To make a fair comparison, these four algorithms are executed on our platform with the same time limit $\hat{t}$, and we execute each algorithm 30 times to solve each instance. Detailed comparative results of FIS and state-of-the-art algorithms on 28 ER\_100 instances and 22 ER\_200 instances are reported in Tables \ref{Tab:Comparisons Among FIS and State-of-the-art Algorithms on 28 ER100 Instances} and \ref{Tab:Comparisons Among FIS and State-of-the-art Algorithms on 22 ER200 Instances}, respectively.

\begin{table*}[!htbp]
    \centering
    \footnotesize
    \caption{Performance Comparison between FIS and State-of-the-art Algorithms on 28 ER\_100 Instances}
    \label{Tab:Comparisons Among FIS and State-of-the-art Algorithms on 28 ER100 Instances}
    \resizebox{\textwidth}{!}{
        \begin{tabular}{lr|rrr|rrr|rrr|rrr}
            \toprule[0.75pt]
            \multicolumn{2}{c}{}         & \multicolumn{3}{c}{GRASP}    & \multicolumn{3}{c}{GRASP/PR}         & \multicolumn{3}{c}{MA}          & \multicolumn{3}{c}{\textbf{FIS}}\\
            \cmidrule[0.5pt](lr){3-5} \cmidrule[0.5pt](lr){6-8} \cmidrule[0.5pt](lr){9-11} \cmidrule[0.5pt](lr){12-14}
            \multicolumn{1}{l}{Instance} & \multicolumn{1}{c}{$\alpha$} & \multicolumn{1}{c}{$\hat{f}$}        & \multicolumn{1}{c}{$\overline{f}$}   & \multicolumn{1}{c}{$\overline{t}$} & \multicolumn{1}{c}{$\hat{f}$} & \multicolumn{1}{c}{$\overline{f}$}   & \multicolumn{1}{c}{$\overline{t}$} & \multicolumn{1}{c}{$\hat{f}$} & \multicolumn{1}{c}{$\overline{f}$}   & \multicolumn{1}{c}{$\overline{t}$} & \multicolumn{1}{c}{$\hat{f}$} & \multicolumn{1}{c}{$\overline{f}$} & \multicolumn{1}{c}{$\overline{t}$} \\
            \midrule[0.5pt]
            ER\_100\_0.05\_0.2\_0        & 0.2                          & \textbf{26}                          & 26.10                                & 7.11                               & \textbf{26}                   & 26.37                                & 8.99                               & \textbf{26}                   & 26.70                                & 0.27                               & \textbf{26}                   & \textbf{26.00}                     & 0.85                               \\
            ER\_100\_0.05\_0.2\_2        & 0.2                          & \textbf{27}                          & \textbf{27.00}                       & 5.08                               & \textbf{27}                   & 27.17                                & 6.69                               & \textbf{27}                   & 27.47                                & 0.11                               & \textbf{27}                   & \textbf{27.00}                     & 0.90                               \\
            ER\_100\_0.05\_0.2\_3        & 0.2                          & \textbf{30}                          & \textbf{30.93}                       & 8.08                               & \textbf{30}                   & 31.40                                & 11.05                              & 31                            & 31.47                                & 0.15                               & 31                            & 31.00                              & 0.98                               \\
            ER\_100\_0.05\_0.2\_4        & 0.2                          & \textbf{28}                          & 28.10                                & 8.22                               & \textbf{28}                   & 28.47                                & 8.19                               & \textbf{28}                   & 28.73                                & 0.12                               & \textbf{28}                   & \textbf{28.00}                     & 1.56                               \\
            ER\_100\_0.05\_0.5\_0        & 0.4                          & 25                                   & 25.00                                & 3.42                               & \textbf{24}                   & \textbf{24.97}                       & 4.50                               & 25                            & 25.87                                & 0.29                               & 25                            & 25.00                              & 3.11                               \\
            ER\_100\_0.05\_0.5\_1        & 0.4                          & \textbf{22}                          & 22.37                                & 9.46                               & \textbf{22}                   & 22.83                                & 8.55                               & \textbf{22}                   & 22.20                                & 0.05                               & \textbf{22}                   & \textbf{22.00}                     & 0.80                               \\
            ER\_100\_0.05\_0.5\_3        & 0.4                          & \textbf{22}                          & \textbf{22.00}                       & 2.46                               & \textbf{22}                   & 22.07                                & 4.67                               & \textbf{22}                   & 22.10                                & 0.04                               & \textbf{22}                   & \textbf{22.00}                     & 0.80                               \\
            ER\_100\_0.05\_0.8\_0        & 0.6                          & \textbf{19}                          & 19.07                                & 7.93                               & \textbf{19}                   & 19.23                                & 9.27                               & \textbf{19}                   & 19.73                                & 1.10                               & \textbf{19}                   & \textbf{19.00}                     & 0.71                               \\
            ER\_100\_0.05\_0.8\_3        & 0.6                          & \textbf{17}                          & 17.17                                & 8.39                               & \textbf{17}                   & 17.77                                & 11.30                              & \textbf{17}                   & 18.20                                & 0.01                               & \textbf{17}                   & \textbf{17.00}                     & 0.60                               \\
            ER\_100\_0.06\_0.2\_3        & 0.2                          & \textbf{35}                          & 35.87                                & 8.51                               & \textbf{35}                   & 36.27                                & 10.76                              & \textbf{35}                   & \textbf{35.33}                       & 4.02                               & \textbf{35}                   & 35.77                              & 2.39                               \\
            ER\_100\_0.06\_0.2\_4        & 0.2                          & \textbf{35}                          & 35.80                                & 6.86                               & \textbf{35}                   & 35.73                                & 11.58                              & \textbf{35}                   & \textbf{35.10}                       & 0.79                               & \textbf{35}                   & 35.40                              & 8.24                               \\
            ER\_100\_0.06\_0.5\_3        & 0.4                          & \textbf{29}                          & 29.27                                & 10.78                              & \textbf{29}                   & 29.47                                & 12.30                              & \textbf{29}                   & 29.77                                & 0.19                               & \textbf{29}                   & \textbf{29.00}                     & 1.11                               \\
            ER\_100\_0.06\_0.8\_4        & 0.6                          & \textbf{20}                          & 20.63                                & 6.89                               & \textbf{20}                   & 21.07                                & 9.37                               & \textbf{20}                   & 20.67                                & 0.63                               & \textbf{20}                   & \textbf{20.00}                     & 0.89                               \\
            ER\_100\_0.07\_0.2\_1        & 0.2                          & \textbf{39}                          & 39.17                                & 8.27                               & \textbf{39}                   & 39.47                                & 10.88                              & \textbf{39}                   & 39.33                                & 0.22                               & \textbf{39}                   & \textbf{39.00}                     & 1.15                               \\
            ER\_100\_0.07\_0.2\_4        & 0.2                          & \textbf{36}                          & 36.17                                & 9.45                               & \textbf{36}                   & 36.67                                & 9.03                               & \textbf{36}                   & 36.37                                & 0.09                               & \textbf{36}                   & \textbf{36.00}                     & 1.24                               \\
            ER\_100\_0.07\_0.5\_0        & 0.4                          & \textbf{34}                          & 34.37                                & 9.57                               & \textbf{34}                   & 34.60                                & 12.85                              & \textbf{34}                   & 34.60                                & 0.11                               & \textbf{34}                   & \textbf{34.00}                     & 1.01                               \\
            ER\_100\_0.07\_0.5\_2        & 0.4                          & \textbf{31}                          & 31.93                                & 8.85                               & \textbf{31}                   & 31.47                                & 12.99                              & \textbf{31}                   & 32.00                                & 0.17                               & \textbf{31}                   & \textbf{31.00}                     & 1.60                               \\
            ER\_100\_0.07\_0.8\_0        & 0.6                          & 25                                   & 25.53                                & 7.20                               & 25                            & 26.33                                & 10.61                              & \textbf{24}                   & 24.90                                & 1.02                               & \textbf{24}                   & \textbf{24.00}                     & 3.99                               \\
            ER\_100\_0.07\_0.8\_4        & 0.6                          & 26                                   & 26.70                                & 4.76                               & 26                            & 27.17                                & 12.39                              & \textbf{25}                   & 25.63                                & 0.87                               & \textbf{25}                   & \textbf{25.07}                     & 5.73                               \\
            ER\_100\_0.08\_0.5\_3        & 0.4                          & \textbf{37}                          & 38.33                                & 8.26                               & 38                            & 39.07                                & 11.25                              & \textbf{37}                   & 38.03                                & 3.75                               & \textbf{37}                   & \textbf{37.67}                     & 7.37                               \\
            ER\_100\_0.08\_0.5\_4        & 0.4                          & 36                                   & 37.13                                & 8.58                               & \textbf{35}                   & 36.97                                & 14.17                              & \textbf{35}                   & 37.00                                & 3.06                               & 36                            & \textbf{36.00}                     & 1.15                               \\
            ER\_100\_0.08\_0.8\_2        & 0.6                          & 29                                   & 30.03                                & 13.08                              & 28                            & 29.70                                & 14.89                              & \textbf{27}                   & 27.87                                & 0.50                               & \textbf{27}                   & \textbf{27.00}                     & 2.84                               \\
            ER\_100\_0.08\_0.8\_4        & 0.6                          & 29                                   & 29.90                                & 5.67                               & 30                            & 30.70                                & 14.57                              & \textbf{28}                   & 29.13                                & 0.87                               & \textbf{28}                   & \textbf{28.00}                     & 8.39                               \\
            ER\_100\_0.09\_0.2\_1        & 0.2                          & 45                                   & 46.20                                & 10.68                              & \textbf{44}                   & 46.40                                & 14.64                              & 45                            & 45.23                                & 0.81                               & 45                            & \textbf{45.00}                     & 2.65                               \\
            ER\_100\_0.09\_0.2\_4        & 0.2                          & \textbf{45}                          & 45.60                                & 12.14                              & \textbf{45}                   & 46.57                                & 9.89                               & \textbf{45}                   & 45.33                                & 0.16                               & \textbf{45}                   & \textbf{45.00}                     & 1.35                               \\
            ER\_100\_0.09\_0.5\_2        & 0.4                          & 40                                   & 40.33                                & 10.31                              & 40                            & 41.10                                & 11.84                              & \textbf{39}                   & 40.40                                & 0.28                               & \textbf{39}                   & \textbf{39.03}                     & 4.46                               \\
            ER\_100\_0.09\_0.8\_1        & 0.6                          & 28                                   & 29.07                                & 9.18                               & 28                            & 29.57                                & 15.95                              & \textbf{27}                   & 27.60                                & 2.49                               & \textbf{27}                   & \textbf{27.00}                     & 4.20                               \\
            ER\_100\_0.09\_0.8\_2        & 0.6                          & \textbf{26}                          & 26.47                                & 10.15                              & \textbf{26}                   & 27.40                                & 14.92                              & \textbf{26}                   & 26.80                                & 0.19                               & \textbf{26}                   & \textbf{26.00}                     & 1.04                               \\
            \midrule[0.5pt]
            \#Wins                       & $-$                          & 1                                    & 1                                    & $-$                                & 4                             & 1                                    & $-$                                & 1                             & 2                                    & $-$                                & $-$                           & $-$                                & $-$                                \\
            \#Ties                       & $-$                          & 21                                   & 3                                    & $-$                                & 17                            & 0                                    & $-$                                & 27                            & 0                                    & $-$                                & $-$                           & $-$                                & $-$                                \\
            \#Losses                     & $-$                          & 6                                    & 24                                   & $-$                                & 7                             & 27                                   & $-$                                & 8                             & 10                                   & $-$                                & $-$                           & $-$                                & $-$                                \\
            \midrule[0.5pt]
            p-value                      & $-$                          & \multicolumn{1}{l}{\textbf{2.9E-02}} & \multicolumn{1}{l}{\textbf{7.3E-06}} & $-$                                & \multicolumn{1}{l}{1.4E-01}   & \multicolumn{1}{l}{\textbf{2.1E-06}} & $-$                                & \multicolumn{1}{l}{8.4E-01}   & \multicolumn{1}{l}{\textbf{7.5E-06}} & $-$                                & $-$                           & $-$                                & $-$                                \\
            \bottomrule[0.75pt]
        \end{tabular}}
\end{table*}

\begin{table*}[!htbp]
    \centering
    \footnotesize
    \caption{Performance Comparison between FIS and State-of-the-art Algorithms on 22 ER\_200 Instances}
    \label{Tab:Comparisons Among FIS and State-of-the-art Algorithms on 22 ER200 Instances}
    \resizebox{\textwidth}{!}{
        \begin{tabular}{lc|rrr|rrr|rrr|rrr}
            \toprule[0.75pt]
            \multicolumn{2}{c}{}         & \multicolumn{3}{c}{GRASP}    & \multicolumn{3}{c}{GRASP/PR}         & \multicolumn{3}{c}{MA}          & \multicolumn{3}{c}{\textbf{FIS}}                                                                                                                                                                                                                                                                                                                                           \\
            \cmidrule[0.5pt](lr){3-5} \cmidrule[0.5pt](lr){6-8} \cmidrule[0.5pt](lr){9-11} \cmidrule[0.5pt](lr){12-14}
            \multicolumn{1}{l}{Instance} & \multicolumn{1}{c}{$\alpha$} & \multicolumn{1}{c}{$\hat{f}$}        & \multicolumn{1}{c}{$\overline{f}$}   & \multicolumn{1}{c}{$\overline{t}$} & \multicolumn{1}{c}{$\hat{f}$}        & \multicolumn{1}{c}{$\overline{f}$}   & \multicolumn{1}{c}{$\overline{t}$} & \multicolumn{1}{c}{$\hat{f}$} & \multicolumn{1}{c}{$\overline{f}$}   & \multicolumn{1}{c}{$\overline{t}$} & \multicolumn{1}{c}{$\hat{f}$} & \multicolumn{1}{c}{$\overline{f}$} & \multicolumn{1}{c}{$\overline{t}$} \\
            \midrule[0.5pt]
            ER\_200\_0.05\_0.2\_0        & 0.2                          & 96                                   & 97.43                                & 221.40                             & 96                                   & 98.97                                & 283.71                             & \textbf{91}                   & \textbf{91.17}                       & 92.52                              & \textbf{91}                   & 92.53                              & 146.66                             \\
            ER\_200\_0.05\_0.2\_1        & 0.2                          & 95                                   & 97.40                                & 215.26                             & 95                                   & 98.80                                & 277.63                             & \textbf{92}                   & \textbf{92.20}                       & 3.86                               & \textbf{92}                   & 92.70                              & 184.85                             \\
            ER\_200\_0.05\_0.2\_3        & 0.2                          & 96                                   & 97.47                                & 207.23                             & 96                                   & 98.63                                & 277.65                             & \textbf{92}                   & \textbf{92.30}                       & 95.46                              & \textbf{92}                   & 92.97                              & 93.10                              \\
            ER\_200\_0.05\_0.5\_0        & 0.4                          & 85                                   & 86.67                                & 193.43                             & 86                                   & 90.30                                & 152.15                             & \textbf{81}                   & 83.40                                & 44.20                              & \textbf{81}                   & \textbf{81.97}                     & 42.97                              \\
            ER\_200\_0.05\_0.5\_4        & 0.4                          & 80                                   & 81.43                                & 219.65                             & 83                                   & 85.60                                & 97.04                              & \textbf{78}                   & 80.00                                & 10.92                              & \textbf{78}                   & \textbf{78.57}                     & 143.39                             \\
            ER\_200\_0.05\_0.8\_3        & 0.6                          & 69                                   & 71.33                                & 140.64                             & 65                                   & 69.17                                & 281.05                             & \textbf{59}                   & 59.17                                & 29.18                              & \textbf{59}                   & \textbf{59.00}                     & 6.94                               \\
            ER\_200\_0.06\_0.5\_1        & 0.4                          & 90                                   & 92.63                                & 201.65                             & 96                                   & 98.17                                & 235.26                             & \textbf{87}                   & 91.07                                & 27.74                              & \textbf{87}                   & \textbf{88.37}                     & 174.98                             \\
            ER\_200\_0.06\_0.5\_2        & 0.4                          & 94                                   & 98.23                                & 217.14                             & 98                                   & 102.10                               & 237.84                             & \textbf{90}                   & 93.23                                & 17.56                              & 91                            & \textbf{91.00}                     & 98.19                              \\
            ER\_200\_0.06\_0.8\_1        & 0.6                          & 76                                   & 76.00                                & 69.65                              & 71                                   & 72.67                                & 309.96                             & \textbf{63}                   & \textbf{63.10}                       & 44.53                              & \textbf{63}                   & 63.17                              & 52.76                              \\
            ER\_200\_0.06\_0.8\_4        & 0.6                          & 73                                   & 73.77                                & 153.37                             & 71                                   & 73.33                                & 247.01                             & \textbf{63}                   & \textbf{63.63}                       & 99.79                              & \textbf{63}                   & 63.93                              & 87.12                              \\
            ER\_200\_0.07\_0.2\_1        & 0.2                          & 112                                  & 113.30                               & 150.06                             & 111                                  & 113.30                               & 207.37                             & \textbf{109}                  & \textbf{109.27}                      & 7.63                               & \textbf{109}                  & 109.33                             & 132.45                             \\
            ER\_200\_0.07\_0.5\_1        & 0.4                          & 102                                  & 104.30                               & 248.59                             & 102                                  & 104.53                               & 280.22                             & \textbf{93}                   & 94.27                                & 9.27                               & \textbf{93}                   & \textbf{93.03}                     & 44.69                              \\
            ER\_200\_0.07\_0.5\_4        & 0.4                          & 100                                  & 103.00                               & 234.56                             & 101                                  & 104.40                               & 257.09                             & \textbf{92}                   & 93.67                                & 31.18                              & \textbf{92}                   & \textbf{92.03}                     & 116.22                             \\
            ER\_200\_0.07\_0.8\_3        & 0.6                          & 79                                   & 79.00                                & 2.24                               & 75                                   & 76.73                                & 257.64                             & \textbf{66}                   & 69.83                                & 179.65                             & \textbf{66}                   & \textbf{66.53}                     & 94.78                              \\
            ER\_200\_0.08\_0.2\_4        & 0.2                          & 116                                  & 117.50                               & 169.02                             & 115                                  & 117.97                               & 315.08                             & \textbf{113}                  & \textbf{113.00}                      & 41.03                              & 114                           & 115.07                             & 176.51                             \\
            ER\_200\_0.08\_0.5\_0        & 0.4                          & 106                                  & 107.87                               & 212.24                             & 107                                  & 108.90                               & 286.06                             & \textbf{98}                   & \textbf{98.50}                       & 65.35                              & \textbf{98}                   & 98.70                              & 153.04                             \\
            ER\_200\_0.08\_0.5\_2        & 0.4                          & 108                                  & 109.37                               & 211.76                             & 105                                  & 107.43                               & 302.11                             & \textbf{96}                   & 96.77                                & 72.23                              & \textbf{96}                   & \textbf{96.07}                     & 158.81                             \\
            ER\_200\_0.08\_0.8\_3        & 0.6                          & 78                                   & 78.20                                & 114.12                             & 76                                   & 77.30                                & 260.92                             & 72                            & 73.07                                & 183.50                             & \textbf{68}                   & \textbf{68.30}                     & 65.05                              \\
            ER\_200\_0.08\_0.8\_4        & 0.6                          & 80                                   & 80.00                                & 0.01                               & 77                                   & 78.20                                & 222.66                             & 73                            & 74.00                                & 162.58                             & \textbf{68}                   & \textbf{69.00}                     & 115.02                             \\
            ER\_200\_0.09\_0.2\_1        & 0.2                          & 121                                  & 123.03                               & 212.57                             & 121                                  & 123.00                               & 332.78                             & \textbf{118}                  & \textbf{118.53}                      & 97.83                              & 119                           & 119.93                             & 151.51                             \\
            ER\_200\_0.09\_0.5\_1        & 0.4                          & 113                                  & 113.60                               & 186.11                             & 108                                  & 111.60                               & 278.33                             & \textbf{100}                  & 100.73                               & 85.09                              & \textbf{100}                  & \textbf{100.57}                    & 133.87                             \\
            ER\_200\_0.09\_0.8\_3        & 0.6                          & 79                                   & 79.00                                & 0.52                               & 77                                   & 78.03                                & 211.68                             & 74                            & 75.00                                & 152.68                             & \textbf{69}                   & \textbf{69.93}                     & 132.36                             \\
            \midrule[0.5pt]
            \#Win                        & $-$                          & 0                                    & 0                                    & $-$                                & 0                                    & 0                                    & $-$                                & 3                             & 9                                    & $-$                                & $-$                           & $-$                                & $-$                                \\
            \#Ties                       & $-$                          & 0                                    & 0                                    & $-$                                & 0                                    & 0                                    & $-$                                & 16                            & 0                                    & $-$                                & $-$                           & $-$                                & $-$                                \\
            \#Losses                     & $-$                          & 22                                   & 22                                   & $-$                                & 22                                   & 22                                   & $-$                                & 3                             & 13                                   & $-$                                & $-$                           & $-$                                & $-$                                \\
            \midrule[0.5pt]
            p-value                      & $-$                          & \multicolumn{1}{l}{\textbf{2.4E-07}} & \multicolumn{1}{l}{\textbf{2.4E-07}} & $-$                                & \multicolumn{1}{l}{\textbf{2.4E-07}} & \multicolumn{1}{l}{\textbf{2.4E-07}} & $-$                                & \multicolumn{1}{l}{1.7E-01}   & \multicolumn{1}{l}{\textbf{2.9E-02}} & $-$                                & $-$                           & $-$                                & $-$                                \\
            \bottomrule[0.75pt]
        \end{tabular}}
\end{table*}

In Tables \ref{Tab:Comparisons Among FIS and State-of-the-art Algorithms on 28 ER100 Instances} and \ref{Tab:Comparisons Among FIS and State-of-the-art Algorithms on 22 ER200 Instances}, columns 1 and 2 present the instance name (Instance) and $\alpha$ value. Columns 3-5 gives the results of GRASP, including the best results found during 30 runs, average result and average time in seconds needed to obtain the best result. Correspondingly, columns 6-8, 9-11, and 12-14, respectively, present the results of GRASP/PR, MA and FIS. Note that the best value of each performance indicator is in bold. In Tables \ref{Tab:Comparisons Among FIS and State-of-the-art Algorithms on 28 ER100 Instances} and \ref{Tab:Comparisons Among FIS and State-of-the-art Algorithms on 22 ER200 Instances}, we provide the number of instances on which FIS's result is better than (\#Wins), equal to (\#Ties), and worse than (\#Loses) the corresponding algorithms'. At its bottom, we give the p-values of the Wilcoxon signed ranks test.

As we can see from Table \ref{Tab:Comparisons Among FIS and State-of-the-art Algorithms on 28 ER100 Instances}, FIS competes very favorably with its three peers on 28 ER\_100 instances, by attaining 7 new upper bounds and matching 19 best-known bounds. Although our FIS show better performance than three-of-the-art algorithms, at a significance level of 0.05, the performance difference among them is statistically marginal in terms of $\hat{f}$. In terms of $\overline{f}$, FIS significantly outperforms GRASP, GRASP/PR and MA.

From Table \ref{Tab:Comparisons Among FIS and State-of-the-art Algorithms on 22 ER200 Instances}, which reports the results of ER\_200 instances, we first observe that FIS finds new upper bounds on all 22 instances. At a significance level of 0.05, FIS significantly outperforms GRASP and GRASP/PR in term of both $\hat{f}$ and $\overline{f}$. Compared to MA, FIS obtains better $\hat{f}$ values on 3 instances, equal $\hat{f}$ values on 16 out of 22 instances, and worse $\hat{f}$ values on remaining 3 instances. However, the performance difference between MA and FIS is not significant. The above observations show that FIS is highly competitive against the state-of-the-art algorithms.

\section{Additional Experimental Results}
\label{Sec:Experimental Result Analysis}

To gain the deeper understanding of FIS, we perform additional experiments. In particular, we conduct five groups of experiments: 1) to perform parameter sensitivity analysis of FIS; 2) to compare the run-time distributions of FIS and MA; 3) to evaluate the benefit of our tabu search strategy; 4) to confirm the superiority of FIR over DBC \cite{Zhou2019}; 5) to verify the effectiveness of our auxiliary function $f'(S)$. Our following experiments are conducted on eight representative instances randomly selected from both ER\_100 and ER\_200.

\subsection{Parameter Sensitivity Analysis}
\label{SubSec:Parameter Sensitivity Analysis}

To find the suitable parameter settings of FIS, we use IRACE \cite{Lopez2016}. For each parameter, it requires some candidate values as input, as shown in the column ``Candidate Values'' of Table \ref{Tab:Parameter Analysis of FIS}. Then, it finds out the best one as ``Final Value'', as shown in the 4th column of Table \ref{Tab:Parameter Analysis of FIS}. We apply the default settings of IRACE and set the total time budget for IRACE to 5000 FIS executions. For each execution, its stopping condition is either (a) The number of generations without improvement exceeds 50, or (b) The algorithm reaches time limit $\hat{t}=30$ seconds for ER\_100 and $\hat{t}=500$ seconds for ER\_200 instances.

\begin{table*}[!htbp]
\centering
\small
\caption{Parameter Analysis of FIS}
\label{Tab:Parameter Analysis of FIS}
\begin{tabular}{lllllc}
\toprule[0.75pt]
Parameter  & Description             & Candidate Values               & Final Value & p-value & Section\\
\midrule[0.5pt]
$\eta$     & Randomized Scale Factor & $\{0.2,0.4,0.6,0.8,1.0\}$      & 0.6         & 0.658   & \ref{SubSec:Solution Construction}                 \\
$\rho$     & Itemset Scale Factor    & $\{0.8,0.85,0.9,0.95,1.0\}$    & 0.95        & 0.626   & \ref{SubSec:Frequent Itemset Recombination}          \\
$\hat{\xi}$  & Maximal Iteration Count & $\{1000,2000,3000,4000,5000\}$ & 2000        & 0.993   &
\ref{SubSec:Tabu Search-based Simulated Annealing} \\
$\gamma$   & Tabu Scale Factor       & $\{0.0,0.2,0.4,0.6,0.8\}$      & 0.2         & 0.727   &
\ref{SubSec:Tabu Search-based Simulated Annealing} \\
$\mu$      & Weight Factor           & $\{0.2,0.4,0.6,0.8,1.0\}$      & 0.6         & 0.654   & \ref{SubSec:Rank-based Population Management}                   \\
\bottomrule[0.75pt]
\end{tabular}
\end{table*}

We use the recommended Friedman test \cite{Demsar2006} to further check whether there is a significant difference between each pair of candidate parameter values in terms of FIS performance. To perform the sensitivity of each parameter, we consider the ``Candidate Values'' for each parameter from Table \ref{Tab:Parameter Analysis of FIS} while fixing other parameters to their ``Final Value''. We run FIS 30 times on each instance with the parameter setting, and then record the average objective value. The p-values are given in the 5th column of Table \ref{Tab:Parameter Analysis of FIS}, which are larger than 0.05. These results confirm that the parameters of FIS present no particular sensitivity at a significance level of 0.05.

\subsection{Time-To-Target Plots of FIS and MA}
\label{SubSec:The Time-To-Target Plots of FIS and MA}

\begin{figure*}[!htbp]
    \centering
    \includegraphics[width=1.5\columnwidth]{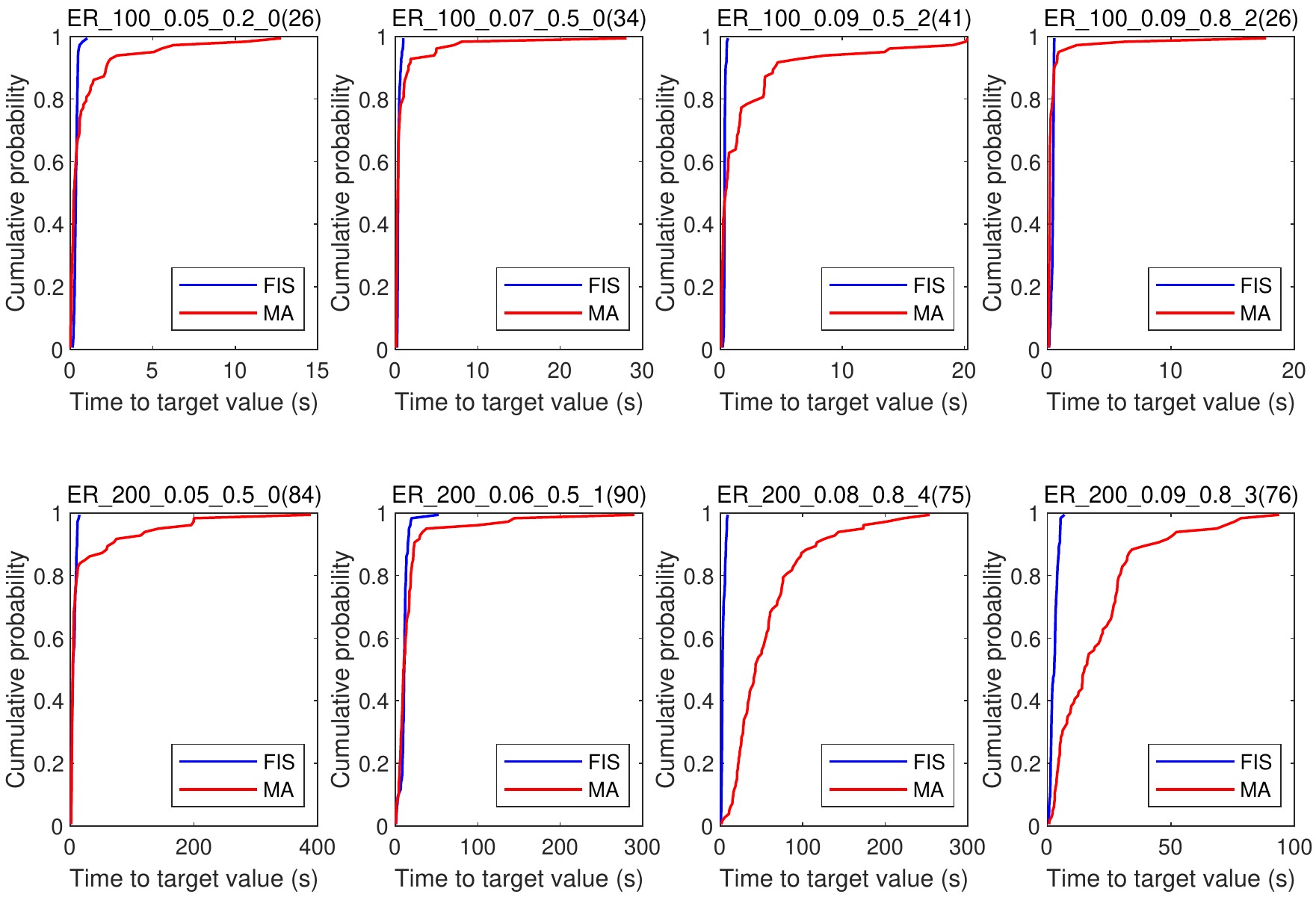}
    \caption{The Run-time Distributions of FIS and MA}
    \label{Fig:TTT Plots}
\end{figure*}

To the best of our knowledge, MA is the best-performing algorithm for $\alpha$-SP in the literature, which is further confirmed in Section \ref{SubSec:Comparative Results Among FIS and State-of-the-art Algorithms}. To gain a deeper insight into the performance of both FIS and MA, we use the time-to-target (TTT) plots \cite{Aiex2007} to analyze their run-time distributions on eight representative instances. A TTT plot is a useful tool to characterize the running times of stochastic algorithms for combinatorial optimization and has been widely used for algorithm comparison. To produce a TTT plot, we run each algorithm 100 times and record the time to obtain a solution at least as good as a given target value. After sorting these times in ascending order, a probability $p_i = \frac{i-0.5}{100}$ is associated with the $t$-th sorted running time $t_i$. These points $(t_i,p_i)$, $i=1,2,\ldots,100$ are then plotted. Figure~\ref{Fig:TTT Plots} shows the TTT plots of FIS and MA on eight instances. Note that Figure~\ref{Fig:TTT Plots} does not provide the TTT plots of both GRASP and GRASP/PR because they are hard to reach the given target values given the limited time.

From Figure~\ref{Fig:TTT Plots}, we can observe that FIS is more likely to find a target solution faster than MA. Considering ER\_100\_0.09\_0.5\_2 as an example, it shows that the probability of finding the target value 41 in at most one second is approximately 60\% for MA, while it is 100\% for FIS. For a larger problem instance ER\_200\_0.09\_0.8\_3, Figure~\ref{Fig:TTT Plots} indicates that the probabilities of reaching the target value 76 in at most 10 seconds are approximately 40\%, and 100\% for MA and FIS, respectively. The analyses of the run-time distributions clearly confirm that FIS outperforms MA.

\subsection{Benefit of Tabu Search Strategy}
\label{SubSec:Benefit of the Tabu Search Strategy}

Our TSSA algorithm combines tabu search strategy with simulated annealing (SA) algorithm. To evaluate its benefit, we experimentally compare TSSA with its a variant TSSA$'$. TSSA$'$ is obtained from TSSA by disabling the latter's tabu search strategy, which is closely related to SA algorithm. Table \ref{Tab:Comparisons between TSSA and its Variant} summarizes the comparative results between TSSA and TSSA$'$ on eight representative instances in terms of the best value, average value, and average time.

\begin{table*}[!hbtp]
    \centering
    \small
    \caption{Comparison between TSSA (with tabu search strategy) and TSSA$'$ (without tabu search strategy)}
    \label{Tab:Comparisons between TSSA and its Variant}
    \begin{threeparttable}
        \setlength{\tabcolsep}{5mm}{
            \begin{tabular}{lc|rrr|rrr}
                \toprule[0.75pt]
                \multicolumn{2}{c}{}         & \multicolumn{3}{c}{TSSA$'$}  & \multicolumn{3}{c}{TSSA}                                                                                                                                                                                          \\
                \cmidrule[0.5pt](lr){3-5}\cmidrule[0.5pt](lr){6-8}
                \multicolumn{1}{l}{Instance} & \multicolumn{1}{c}{$\alpha$} & \multicolumn{1}{c}{$\hat{f}$} & \multicolumn{1}{c}{$\overline{f}$} & \multicolumn{1}{c}{$\overline{t}$} & \multicolumn{1}{c}{$\hat{f}$} & \multicolumn{1}{c}{$\overline{f}$} & \multicolumn{1}{c}{$\overline{t}$} \\
                \midrule[0.5pt]
                ER\_100\_0.05\_0.2\_0        & 0.2                          & \textbf{26}                   & \textbf{26.00}                     & 0.99                               & \textbf{26}                   & \textbf{26.00}                     & 0.49                               \\
                ER\_100\_0.07\_0.5\_0        & 0.4                          & \textbf{34}                   & \textbf{34.00}                     & 0.67                               & \textbf{34}                   & \textbf{34.00}                     & 0.76                               \\
                ER\_100\_0.09\_0.5\_2        & 0.4                          & \textbf{39}                   & \textbf{39.00}                     & 3.64                               & \textbf{39}                   & \textbf{39.00}                     & 3.66                               \\
                ER\_100\_0.09\_0.8\_2        & 0.6                          & \textbf{26}                   & \textbf{26.00}                     & 0.22                               & \textbf{26}                   & \textbf{26.00}                     & 0.16                               \\
                ER\_200\_0.05\_0.5\_0        & 0.4                          & \textbf{82}                   & \textbf{82.93}                     & 176.80                             & \textbf{82}                   & 83.03                              & 147.80                             \\
                ER\_200\_0.06\_0.5\_1        & 0.4                          & 89                            & 89.26                              & 140.00                             & \textbf{88}                   & \textbf{89.10}                     & 178.04                             \\
                ER\_200\_0.08\_0.8\_4        & 0.6                          & 73                            & 73.22                              & 204.09                             & \textbf{72}                   & \textbf{73.10}                     & 196.75                             \\
                ER\_200\_0.09\_0.8\_3        & 0.6                          & 74                            & 74.44                              & 160.35                             & \textbf{73}                   & \textbf{74.30}                     & 216.28                             \\
                \midrule[0.5pt]
                avg.rank                     & $-$                          & 1.69                          & 1.63                               & $-$                                & 1.31                          & 1.38                               & $-$                                \\
                \bottomrule[0.75pt]
            \end{tabular}}
    \end{threeparttable}
\end{table*}

From Table \ref{Tab:Comparisons between TSSA and its Variant}, we observe that TSSA has excellent performance. It achieves smaller average ranks in terms of both $\hat{f}$ and $\overline{f}$. In particular, it can find better results on 3 out of 8 instances, and the same results on remaining 5 instances in terms of $\hat{f}$. In terms of $\overline{f}$, TSSA finds better results on 3 out of 8 instances, and the same results on 4 instances, and the worse result on one instance. These observations prove the effectiveness of the tabu search strategy in TSSA.

\subsection{Superiority of Frequent Itemset Recombination}
\label{SubSec:Superiority of the Proposed Frequent Itemset Recombination}

In this work, we employ FIR to construct offspring solutions. To demonstrate its superiority over existing crossover, we experimentally compare it with its an alternative version FIS$'$. FIS$'$ is obtained from FIS by replacing FIR with a double backbone crossover (DBC) originally proposed in \cite{Zhou2019}. DBC treats the common nodes shared by two parents as the first backbone and directly inherits them to form a partial solution. Then it adds some nodes belonging to only one parent (i.e., the second backbone) into the partial solution in a probabilistic way. Finally, it repairs the partial solution until a feasible solution is obtained.

\begin{figure}[!htbp]
    \centering
    \includegraphics[width=1.0\columnwidth]{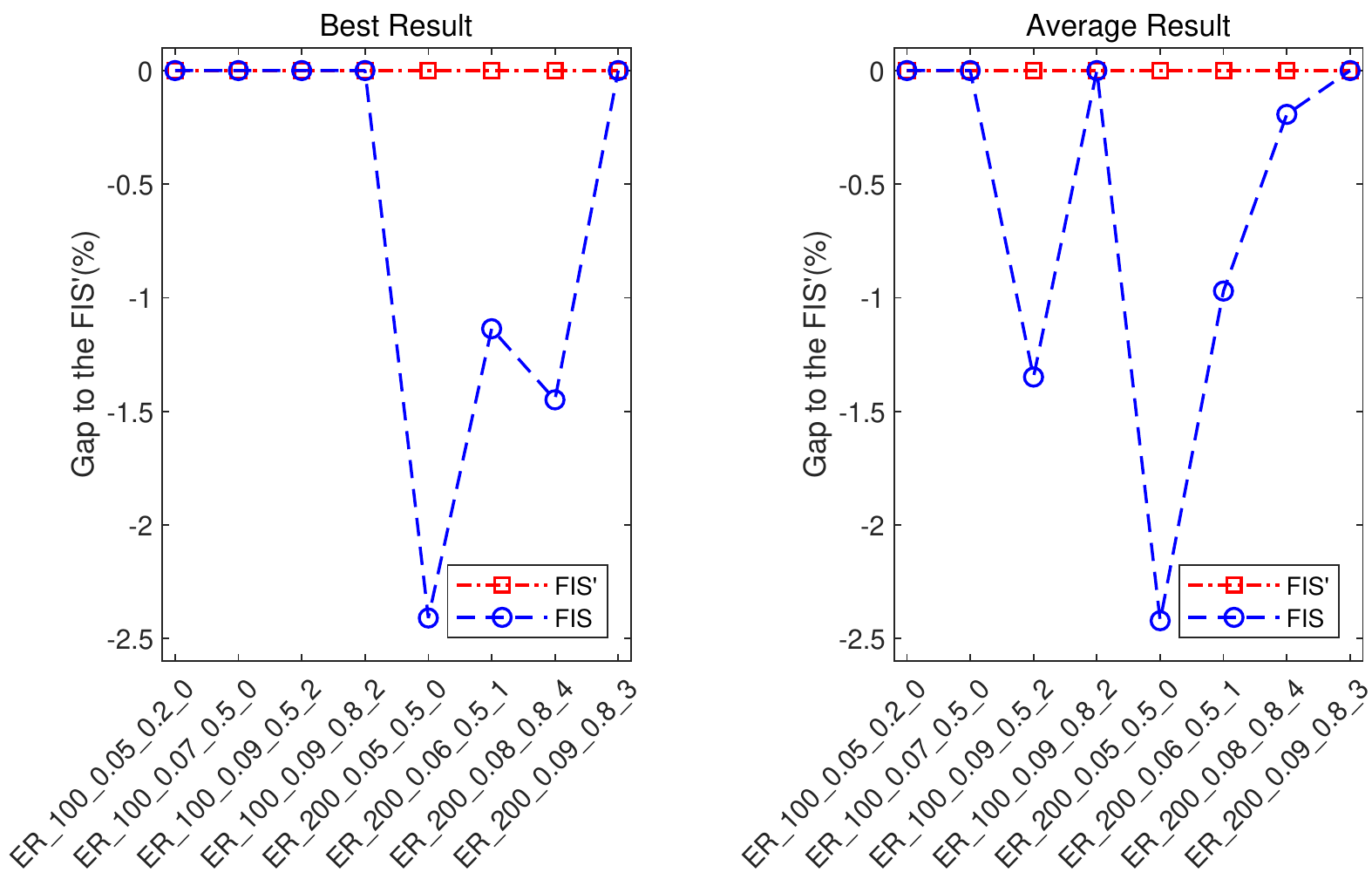}
    \caption{Comparison between FIS (with frequent itemset recombination) and FIS$'$ (with double backbone crossover)}
    \label{Fig:Comparisons between FIS and Its Variant}
\end{figure}

Comparative results of FIS and FIS$'$ on eight representative instances in terms of the best and average values are shown in the left and right of Figure~\ref{Fig:Comparisons between FIS and Its Variant}, respectively. The $x$-axis indicates instances, and $y$-axis provides performance gaps. By treating FIS$'$ as a baseline algorithm, we calculate their performance gap as ${(f-\tilde{f})}/{\tilde{f}} \times 100\%$, where $f$ is its result and $\tilde{f}$ is the result of FIS$'$. A gap smaller than zero means that FIS achieves a better result on the corresponding instance.

From Figure~\ref{Fig:Comparisons between FIS and Its Variant}, we observe that FIS can attain a better $\hat{f}$ value for all eight instances including three new improved results (see the values that below 0 in the left of Figure~\ref{Fig:Comparisons between FIS and Its Variant}). FIS also demonstrates better performance than FIS$'$ in terms of average result. In particular, FIS outperforms FIS$'$ on 4 out of 8 instances, and achieves the same result on remaining four instances, as shown in the right of Figure~\ref{Fig:Comparisons between FIS and Its Variant}. These results confirm the superiority of FIR over DBC.

\subsection{Effectiveness of Proposed Auxiliary Function}
\label{SubSec:Superiority of Our Auxiliary Function}

Given a candidate solution $S$ of $K$-decision $\alpha$-SP, we first use an auxiliary function $f'(S)$ to evaluate it, as defined in (\ref{Equ:Auxiliary Function}). $f'(S)$ counts the number of nodes in the largest connected component that has over $\lceil \alpha \cdot n \rceil$ nodes. To demonstrate its effectiveness, we compare it to two auxiliary functions: $f''(S)$ and $f'''(S)$. The former was original proposed in \cite{Zhou2019}, i.e.,
\begin{equation} \label{Equ:Auxiliary Function 1}
    f''(S) = \sum_{i=1}^{T} \max(|\mathcal{C}_i|-\lceil \alpha \cdot n \rceil, 0)
\end{equation}
which counts the total number of nodes in all $T$ connected components that have over $\lceil \alpha \cdot n \rceil$ nodes in the residual graph. The latter computes the number of connected components with more than $\lceil \alpha \cdot n \rceil$ nodes after removing $S$, i.e.,
\begin{equation} \label{Equ:Auxiliary Function 0}
    f'''(S) = \sum_{i=1}^{T} x_i
\end{equation}
where the binary variable $x_i=1~(i=1,2,\ldots,T)$ if the size of the $i$-th connected component $\mathcal{C}_i$ is larger than $\lceil \alpha \cdot n \rceil$, i.e., $|\mathcal{C}_i| > \lceil \alpha \cdot n \rceil$, and $x_i=0$ otherwise. Note that both $f'(S)$ and $f'''(S)$ can be considered as a special case of $f''(S)$. Importantly, $f'(S)$ only considers the largest connected component instead of all $T$ connected components in the residual graph. Thus it is faster to compute than $f''(S)$.

Table \ref{Tab:Comparisons among FIS Algorithms with Three Different Auxiliary Functions} summarizes the comparative results of FIS with three different auxiliary functions. At its bottom, we provide the average rank of both best and average values. It is seen that FIS with $f'(S)$ obtains the smallest average rank in terms of both $\hat{f}$ and $\overline{f}$. In particular, FIS with $f'(S)$ achieves better results than FIS with $f'''(S)$ on all 8 instances in terms of both $\hat{f}$ and $\overline{f}$. Compared to FIS with $f''(S)$, FIS with $f'(S)$ finds better results on 2 out of 8 instances and the same results on remaining 6 instances in terms of $\hat{f}$. In terms of $\overline{f}$, FIS with $f'(S)$ finds better or the same results on 6 out of 8 instances, and slightly worse results on two remaining instances. Moreover, FIS with $f'(S)$ achieves these better results in shorter computation time. These results confirm the effectiveness of $f'(S)$.

\begin{table*}[!hbt]
    \centering
    \small
    \caption{Comparison among FIS Algorithms with Three Different Auxiliary Functions}
    \label{Tab:Comparisons among FIS Algorithms with Three Different Auxiliary Functions}
    \begin{threeparttable}
        \begin{tabular}{lc|rrr|rrr|rrr}
            \toprule[0.75pt]
            \multicolumn{2}{c}{}         & \multicolumn{3}{c}{$f'''(S)$} & \multicolumn{3}{c}{$f''(S)$}  & \multicolumn{3}{c}{\textbf{$f'(S)$}}                                                                                                                                                                                                                                                          \\
            \cmidrule[0.5pt](lr){3-5} \cmidrule[0.5pt](lr){6-8} \cmidrule[0.5pt](lr){9-11}
            \multicolumn{1}{l}{Instance} & \multicolumn{1}{c}{$\alpha$}  & \multicolumn{1}{c}{$\hat{f}$} & \multicolumn{1}{c}{$\overline{f}$}   & \multicolumn{1}{c}{$\overline{t}$} & \multicolumn{1}{c}{$\hat{f}$} & \multicolumn{1}{c}{$\overline{f}$} & \multicolumn{1}{c}{$\overline{t}$} & \multicolumn{1}{c}{$\hat{f}$} & \multicolumn{1}{c}{$\overline{f}$} & \multicolumn{1}{c}{$\overline{t}$} \\
            \midrule[0.5pt]
            ER\_100\_0.05\_0.2\_0        & 0.2                           & 30                            & 32.20                                & 2.26                               & \textbf{26}                   & 26.03                              & 0.26                               & \textbf{26}                   & \textbf{26.00}                     & 2.64                               \\
            ER\_100\_0.07\_0.5\_0        & 0.4                           & 39                            & 42.73                                & 2.77                               & \textbf{34}                   & \textbf{34.00}                     & 0.35                               & \textbf{34}                   & \textbf{34.00}                     & 0.63                               \\
            ER\_100\_0.09\_0.5\_2        & 0.4                           & 43                            & 46.30                                & 1.37                               & \textbf{39}                   & \textbf{39.00}                     & 2.46                               & \textbf{39}                   & \textbf{39.00}                     & 4.64                               \\
            ER\_100\_0.09\_0.8\_2        & 0.6                           & 30                            & 31.93                                & 4.77                               & \textbf{26}                   & \textbf{26.00}                     & 0.39                               & \textbf{26}                   & \textbf{26.00}                     & 0.58                               \\
            ER\_200\_0.05\_0.5\_0        & 0.4                           & 101                           & 104.47                               & 42.53                              & 82                            & 82.00                              & 26.02                              & \textbf{81}                   & \textbf{81.93}                     & 66.19                              \\
            ER\_200\_0.06\_0.5\_1        & 0.4                           & 106                           & 108.70                               & 37.76                              & \textbf{87}                   & \textbf{88.23}                     & 166.61                             & \textbf{87}                   & 88.43                              & 124.00                             \\
            ER\_200\_0.08\_0.8\_4        & 0.6                           & 77                            & 77.97                                & 47.49                              & 69                            & 69.20                              & 66.90                              & \textbf{68}                   & \textbf{69.00}                     & 88.50                              \\
            ER\_200\_0.09\_0.8\_3        & 0.6                           & 77                            & 77.90                                & 45.28                              & \textbf{69}                   & \textbf{69.80}                     & 130.58                             & \textbf{69}                   & 70.00                              & 147.17                             \\
            \midrule[0.5pt]
            avg.rank                     & $-$                           & 3.00                          & 3.00                                 & $-$                                & 1.63                          & 1.56                               & $-$                                & \textbf{1.37}                 & \textbf{1.44}                      & $-$                                \\
            \bottomrule[0.75pt]
        \end{tabular}
    \end{threeparttable}
\end{table*}

\section{Conclusion}
\label{Sec:Conclusion}

In this paper, we present a frequent itemset-driven search (FIS) to solve an $\alpha$-separator problem ($\alpha$-SP). It integrates the concept of frequent itemset into the well-known memetic algorithm framework, which distinguishes itself from existing memetic algorithms in three aspects: 1) An iterative solution strategy by considering $\alpha$-SP from the viewpoint of constraint satisfaction and then solving a series of $K$-decision $\alpha$-SPs; 2) A frequent itemset recombination operator which employs frequent itemsets among high-quality solutions to construct offspring solutions; and 3) A tabu search-based simulated annealing integrates the tabu strategy into the classic simulated annealing to perform effective local optimization.

Extensive experimental results on 50 widely used benchmark instances have demonstrated the superiority of FIS over three state-of-the-art algorithms (i.e., GRASP, GRASP/PR and MA). In particular, FIS finds improved best results for 29 instances, and matches best known results on 18 instances. Our experimental analyses have confirmed the benefit of both frequent itemset recombination and tabu search-based simulated annealing. For the ablation studies of other modules, we plan to leave them as future work due to space limit. As future work, the following two potential research directions can be pursued. To further improve FIS, it is worth studying some advanced population management strategies to reinforce population diversity and improve the search. In addition, it is interesting to investigate its effectiveness in solving other subset selection problems, e.g., diversity and dispersion problems.


\ifCLASSOPTIONcaptionsoff
  \newpage
\fi

\bibliographystyle{IEEEtran}
\bibliography{mybibfiles}

\end{document}